\documentclass{article}

\usepackage[final,main]{neurips_2026}

\usepackage[utf8]{inputenc}
\usepackage[T1]{fontenc}
\usepackage{hyperref}
\usepackage{url}
\usepackage{xurl}   
\usepackage{booktabs}
\usepackage{array}   
\usepackage{enumitem}
\usepackage{amsmath}
\usepackage{amsfonts}
\usepackage{mathtools}
\usepackage{nicefrac}
\usepackage{microtype}
\usepackage{xcolor}
\usepackage{graphicx}
\usepackage{subcaption}
\usepackage{multirow}
\usepackage{wrapfig}
\usepackage[most]{tcolorbox}
\usepackage{fvextra}   

\definecolor{cSlate}{HTML}{3d5a80}
\definecolor{cRust}{HTML}{c4553a}
\definecolor{cNeu}{HTML}{8a8478}
\definecolor{cBg}{HTML}{f0eeeb}
\definecolor{cInk}{HTML}{2d2926}
\hypersetup{colorlinks=true, linkcolor=cSlate, citecolor=cSlate, urlcolor=cSlate}

\newtcolorbox{logcard}{
  enhanced, breakable,
  arc=5pt, boxrule=0.4pt,
  colback=cBg, colframe=cNeu!50,
  left=8pt, right=8pt, top=6pt, bottom=6pt,
  before skip=6pt, after skip=6pt,
}

\newtcolorbox{promptcard}[1]{
  enhanced, breakable,
  arc=5pt, boxrule=0.4pt,
  colback=cBg, colframe=cNeu!50,
  colbacktitle=cNeu!22, coltitle=cInk,
  fonttitle=\bfseries\small, title={#1},
  left=8pt, right=8pt, top=6pt, bottom=6pt,
  toptitle=3pt, bottomtitle=3pt,
  before skip=8pt, after skip=8pt,
}

\definecolor{cSage}{HTML}{7a9e7e}
\definecolor{cOchre}{HTML}{c49a3a}
\newtcolorbox{roomlog}[1]{
  enhanced, breakable,
  arc=5pt, boxrule=0.4pt,
  colback=white, colframe=cSlate!35,
  colbacktitle=cSlate!12, coltitle=cInk,
  fonttitle=\bfseries\small, title={#1},
  left=8pt, right=8pt, top=4pt, bottom=6pt,
  toptitle=3pt, bottomtitle=3pt,
  before skip=8pt, after skip=8pt,
}
\newcommand{\logmsg}[3]{
  \par\medskip\noindent\textcolor{#1}{\textbf{\small #2}}\hfill{\footnotesize\color{cNeu}$t{=}#3$\,s}\par\nopagebreak\vspace{1pt}\noindent}

\title{AgentRoom: Concurrent Multi-Agent Coding\\in a CRDT-Backed Shared Workspace}

\author{%
  Seonglae Cho\thanks{Correspondence: \texttt{sungle3737@gmail.com}} \\
  Holistic AI \\
  \And
  Donghyun Lee \\
  University of California, Berkeley \\
}

\def\seongland{\raisebox{-1.5pt}{\includegraphics[height=1.05em]{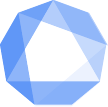}}}
\newcommand{\articlelink}{https://seongland.com/article/agentroom}

\begin{document}

\maketitle

\vspace{-2.5em}
\begin{center}
\begin{tabular}{rl}
\seongland & \url{\articlelink}\\
\end{tabular}
\end{center}

\begin{abstract}
Concurrent \mbox{multi-agent}\ coding offers a division of labor across modules, robustness through redundancy and parallel exploration at the natural granularity of \mbox{multi-file}\ projects.
Realtime collaborative editing protocols solve this coordination problem for human teams with Conflict-free Replicated Data Types (CRDTs), but the LLMs underneath generate one token at a time, and existing \mbox{multi-agent}\ coding systems inherit this serial limit, since they either sequence agents through phase handoffs or pool independent samples without coordination, and a single agent abandons up to half of hard tasks with a \mbox{one-file}\ \mbox{stub-and-exit}.
\textit{AgentRoom} is a realtime collaborative editing protocol for concurrent coding agents.
Its runtime layer exposes \mbox{file-level}\ claim, status and broadcast as MCP tools on a \mbox{CRDT-merged}\ shared filesystem.
Five frontier \mbox{coding-CLI}\ models ran four backend coding tasks, with \mbox{cross-language}\ checks in Python DevBench and Rust+axum.
For \mbox{CLI-stable}\ models, AgentRoom with 2 agents abandons fewer tasks than Solo and varies less from run to run.
At matched compute, one positive mean \mbox{LLM-judge}\ contrast puts AgentRoom over \mbox{parallel-merge}.\ The other contrast, a bundle probe, puts full AgentRoom above each partial case: an ordering rather than a percentage split.
Coordination, not parallelism or \mbox{CRDT-merge},\ is the key design choice.

\end{abstract}

\section{Introduction}

Multi-agent coding systems~\citep{chatdev,metagpt,agilecoder,agentcoder,selforganized,codecor,codesim,semag,multiagentbench,autogen,openhands,generative_agents,agentverse,reflexion,multiagent_runtime_debug,agentsnet,process_models_multiagent} predominantly use sequential turn-taking: design $\to$ implement $\to$ review, each agent waiting for the previous.
Implicit concurrent coordination via CRDTs has been explored~\citep{codecrdt}, with mixed results (21\% speedup / 39\% slowdown across tasks).
Whether \emph{explicit} coordination on a CRDT-merged shared workspace can outperform both the sequential and the implicit-CRDT alternatives at matched compute is the question we address.

\textbf{System.} AgentRoom is a \emph{state-management room} for concurrent coding agents: a structured runtime object with file-level claim semantics, an append-only broadcast log, and each agent's status, exposed to agents as a small set of Model Context Protocol (MCP) tools (\texttt{room\_claim}, \texttt{room\_release}, \texttt{room\_state}, \texttt{room\_broadcast}, \texttt{room\_read}) on top of a CRDT-merged shared filesystem~\citep{shapiro_crdt,yjs,automerge}.
\textbf{Comparison.}
AgentRoom uses a different coordination paradigm from CodeCRDT~\citep{codecrdt}, which infers coordination implicitly and pre-assigns outliner and implementer roles: it replaces both with an explicit room and an advisory protocol (Section~\ref{sec:protocol}), without role pre-assignment or orchestrated handoffs.
We make this comparison at matched compute in Section~\ref{sec:results}: the agents stay the same, but we combine them in different ways.

\begin{wrapfigure}{r}{0.52\linewidth}
\vspace{-1.0em}
\centering
\includegraphics[width=\linewidth]{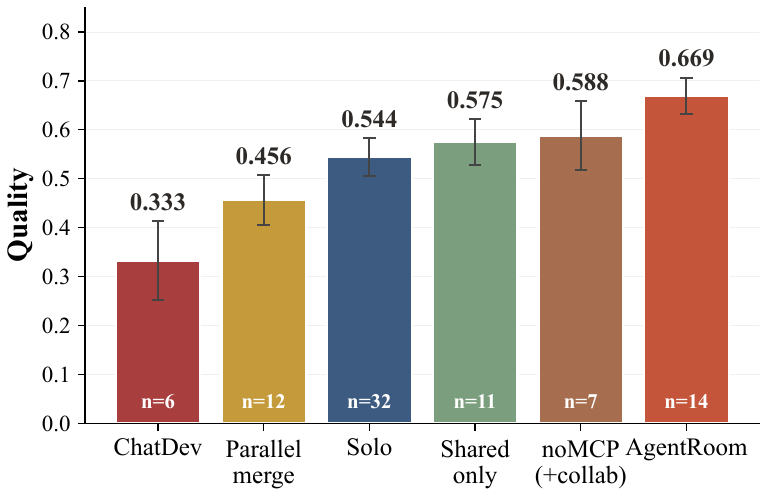}
\caption{T4 six-condition coordination ablation: AgentRoom is the top condition, while concurrency without coordination (parallel-merge) and sequential hand-off (ChatDev-style) fall below the single-agent baseline.}
\label{fig:overview}
\vspace{-1.0em}
\end{wrapfigure}
\textbf{Finding.} We find that two agents in an \mbox{AgentRoom} \emph{suppress the lone-agent ``stub-and-exit'' failure mode} (the agent emits a one-file source skeleton and exits early, judging the task too hard). Across all $12$ model$\times$task strata, the pooled Cochran-Mantel-Haenszel (CMH) estimate puts Solo's odds of abandonment at $\mathbf{13.7}$ times AgentRoom's ($95\%$ CI $[3.9,\,48]$, $\mathbf{p<10^{-5}}$, Tarone homogeneity $p{=}0.92$). Solo's continuity-corrected abandonment odds exceed AgentRoom's in every stratum with events. Figure~\ref{fig:forest} (Appendix~\ref{app:forest}) breaks the results down by model and task.
\textbf{Contrasts.}
At matched compute, AgentRoom leads parallel-merge by $+0.213$ in mean quality on the continuous LLM-judge composite (Welch $p{=}0.003$, both Sonnet, structurally bias-immune; Figure~\ref{fig:overview}). A bundle component probe (Appendix~\ref{app:bundle_probe}) is consistent with the mean ordering shared-only $<$ prompt-only $<$ AgentRoom, with the larger observed step at the MCP coordination layer; at $n{=}7$ the interval on that step spans zero, so we read the split only as an ordering.

\section{Method}

\subsection{AgentRoom: a CRDT-backed shared workspace}

\begin{wrapfigure}[16]{r}{0.50\linewidth}
\vspace{-1.0em}
\centering
\includegraphics[width=\linewidth]{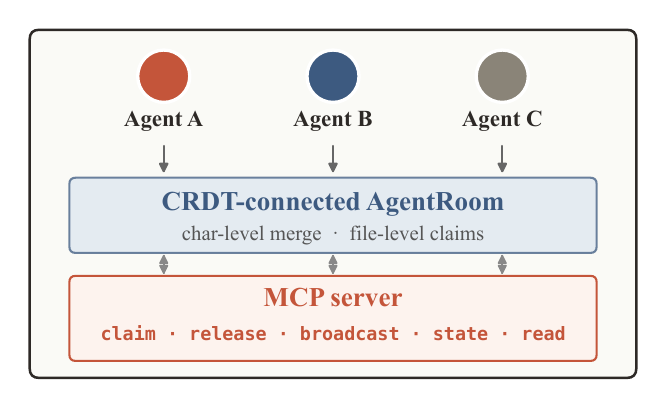}
\caption{AgentRoom architecture: $N$ coding agents share a workspace mediated by a CRDT; the MCP server exposes \texttt{claim}, \texttt{release}, \texttt{broadcast}, \texttt{state}, \texttt{read}.}
\label{fig:architecture}
\vspace{-1.0em}
\end{wrapfigure}
AgentRoom is a single integrated primitive (Figure~\ref{fig:architecture}): a CRDT-mediated shared filesystem $\mathcal{W}$ with $N$ concurrent agents plus a coordination interface.
We co-designed the three components below, and we ablate them only partially (Appendix~\ref{app:bundle_probe}).

\begin{figure}[t]
\centering
\includegraphics[width=\linewidth]{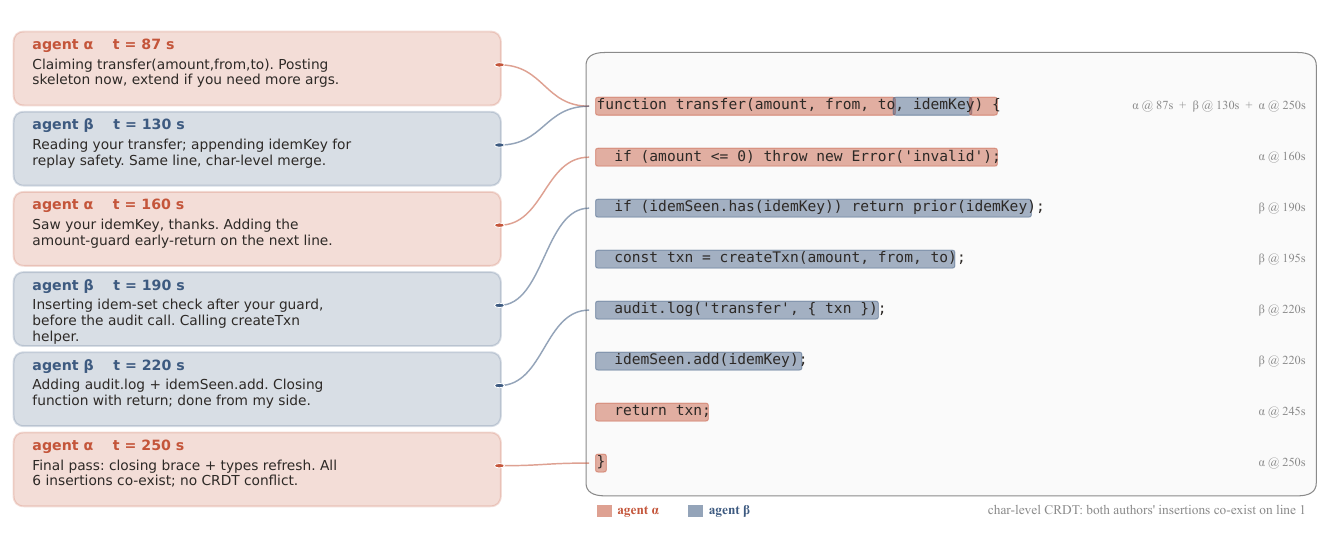}
\caption{Real-time CRDT merge during a single T4 Sonnet AgentRoom~$\times 2$ run. The MCP-mediated chat (left) drives concurrent character-level edits in the shared file (right); both $\alpha$'s and $\beta$'s insertions co-exist on the same line via the Yrs CRDT, with no byte-level conflict raised in this run. Per-message Bezier leaders link each chat broadcast to the code line it talks about.}
\label{fig:crdt_merge}
\end{figure}

\textbf{Shared Workspace + CRDT.} All $N$ agents execute in $\mathcal{W}$, and any file one agent writes is immediately visible to the others.
\texttt{pycrdt} (the Yrs port~\citep{yjs} of Y.js) merges concurrent writes within $\Delta_{\text{crdt}}{=}2$\,s at the character level, and we add a brace/paren-balance sanity check (Figure~\ref{fig:crdt_merge}).

\textbf{Coordination Interface.} The room lets an agent claim a file, broadcast a note and query who holds what through MCP server tools: \texttt{room\_claim(path)} atomically assigns file ownership and rejects if another agent already holds the path; \texttt{room\_broadcast/read} maintain an append-only message log in JSONL; \texttt{room\_state} exposes peer state; the interface enforces ownership at the state-management layer rather than relying on agents to parse text claims. This is the key difference from chat-only multi-agent coordination~(\S\ref{sec:chat_vs_state}).

\subsection{Formal model: collision rate and coordination layer}
\label{sec:formal_model}

We write $\mathcal{A} = \{a_1, \ldots, a_N\}$ for the agents, $\mathcal{F}$ for the file set, and $\mathcal{W}_t$ for the workspace state at time $t$.

\spaceskip=\fontdimen2\font plus \fontdimen3\font minus 0.05\fontdimen4
\paragraph{What SEC does and does not buy.} An op-based sequence CRDT~\citep{shapiro_crdt,rga,json_crdt} underlies the substrate, stamping edits with Lamport timestamps~\citep{lamport_clocks} and satisfying \emph{strong eventual consistency} (SEC); SEC eliminates data loss under concurrent edits, but it is a property of how bytes merge, not of semantic compatibility: if two agents insert incompatible function signatures at the same character offset, both survive in the merged file and break compilation downstream.
\par\spaceskip=0pt\relax

Appendix~\ref{app:algebra} defines edits, patch sequences and edit sets, and states SEC.

\textbf{Coordination layer.} Let $\mathcal{C}_t \subseteq \mathcal{F} \times \mathcal{A}$ be the active claims and $\mathcal{L}_t$ the append-only message log. We model three of the room's MCP tools: \texttt{claim}$(f)$ atomically maps $\mathcal{A}{\times}\mathcal{F}{\to}\{\textsc{ok},\textsc{conflict}\}$, \texttt{broadcast}$(m)$ appends $m$ to $\mathcal{L}_t$, and \texttt{state}$()$ returns $(\mathcal{C}_t,\mathcal{L}_{[t-\tau,t]})$ where $\tau$ is a recent-log time window.
\texttt{claim} is atomic and provides \emph{mutual exclusion under cooperation}: $\forall f,\,\forall t:\;\big|\{i : (f, i) \in \mathcal{C}_t\}\big| \le 1$. As in the advisory locks of Chubby~\citep{chubby}, we enforce claims at the prompt layer rather than in the kernel; SEC still keeps the bytes of an agent that writes without claiming, and the room log shows the violation, as in the cross-agent bug fixes of Appendix~\ref{app:room-logs}.

\textbf{Why the substrate alone is not enough.} Even under pairwise-independent per-agent policies (this gives a lower bound, since bias from the shared task and workspace makes the true rate higher), the probability that any two of the $N$ agents target the same file from a pool of $K_t$ candidates is
\begin{equation}
p_{\text{collision}}(N, K_t) \;=\; 1 - \prod_{k=0}^{N-1}\!\Big(1 - \tfrac{k}{K_t}\Big),
\label{eq:collision}
\end{equation}
which comes to ${\approx}0.20$ at our $N{=}2$, $K_t{\approx}5$ operating point. Colliding edits typically break compilation or tests, and the \texttt{claim} primitive drives this rate to zero at a cost of $O(1)$ per edit.

\textbf{Information density (motivational).} Appendix~\ref{app:infodensity} gives an information-theoretic motivation for why coordinated concurrent edits should beat sequential edits at fixed compute: if we treat patches as random variables over the rubric-coverage space, the per-step joint information across agents scales near-linearly in $N$ when claims keep the joint entropy from collapsing. We present this as motivation for the bundle component probe rather than a formal proof. It is per-step motivation only, not a claim that end-task quality scales in $N$; the demonstrated operating point is $N{=}2$, and quality declines beyond it (\S\ref{sec:scaling}).

\subsection{Collaboration Protocol}
\label{sec:protocol}

\paragraph{Workflow.} Agents receive an advisory six-step workflow: (1) \emph{read} room state before writing; (2) \emph{claim} each file they plan to create; (3) on a claim conflict, \emph{revise} and pick a different file; (4) \emph{write} the claimed files; (5) \emph{poll} for updates between subtasks; (6) \emph{report} completion by broadcast.
\paragraph{Rules in practice.} The protocol has two standing rules. The room rejects claims for files already held by another agent. The prompt tells agents to read existing files before writing and to import from teammates' code. The protocol is advisory at the prompt layer; agents typically deviate when a teammate stalls, an informative pattern documented in Appendix~\ref{app:room-logs}.
Despite this imperfect adherence, we find that AgentRoom outperforms the no-coordination condition.

\subsection{Baselines}

We compare AgentRoom against three alternatives: \emph{Solo} (a single agent at the same wall-clock budget) and, at matched compute, \emph{Shared-only} ($N$ agents on one shared CRDT workspace, no collaboration prompt, no MCP tools) and \emph{Parallel-merge} ($N$ agents in separate workspaces, post-hoc file union, later-timestamp tie-break).

\subsection{Quality Score and Emergence Score}
\label{sec:scoring}

\paragraph{Scorers.} Each run receives a quality composite in $[0,1]$. We use three scorers. Their definitions overlap but differ, and Appendix~\ref{app:scorer} lists their full weights and dimensions. An LLM-judge scores the T4 budget-fair body cells: Sonnet~4.6 with a fixed rubric over four dimensions (spec coverage, correctness signals, code quality, test rigor) with weights $0.35/0.30/0.20/0.15$. For cross-validation we also compute a regex scorer (five dimensions including a test-pass-rate term) and an AST scorer (four dimensions over the TypeScript compiler API). For the cross-model headline ES (Table~\ref{tab:crossmodel}) we use the AST scorer, because ceremonial defensive code can game lexical pattern-matching; for the matched-compute ablation (Table~\ref{tab:ablation}) we use the LLM-judge composite, and the regex/AST composites in Appendix~\ref{app:scorer} preserve the same condition ordering.

\paragraph{Cross-validation.} The LLM-judge correlates with the regex scorer at Pearson $r{=}0.67$ over the T4 runs both scorers see ($n{=}246$, App.~\ref{app:scoring}); regex and AST correlate at Pearson $r{=}0.79$ and Spearman $\rho{=}0.62$ on a cross-check of 30 runs.
Independently of the LLM judge, a deterministic 1-file abandonment classifier decides the Tier I CMH binary outcome (\S\ref{sec:variance}): for runs whose regex-scorer composite is below $0.3$, it reads the run duration and the file count. Our main directional findings (CMH OR$=13.7$, AgentRoom $>$ parallel-merge, AgentRoom $>$ ChatDev) hold under the regex and AST scorers too (Appendix~\ref{app:scoring}).

\paragraph{Emergence score.} The Emergence Score divides the mean quality of $N$ agents by the mean quality of a solo agent at the same model and task, $\mathrm{ES} = \bar s_N/\bar s_1$, where $\mathrm{ES}{>}1$ indicates multi-agent improvement.
For headline cells we report 95\% bootstrap CIs~\citep{efron1979} on ES (10\,000 resamples).
Unless stated otherwise, $N{=}2$.

\paragraph{Judge sensitivity.} We test the primary Sonnet judge for same-family preference bias~\citep{preference_leakage,self_preference} by scoring a subset of $30$ T4 runs again with two cross-judges: Claude-Haiku (Anthropic cross-model) and Codex GPT-5.4 (cross-vendor). Sonnet tracks Haiku at $r{=}0.87$ ($n{=}30$), Sonnet tracks Codex at $r{=}0.86$ ($n{=}26$), and Haiku tracks Codex at $r{=}0.78$ ($n{=}26$). The cross-vendor Codex judge is uniformly stricter, by $0.151$ in mean composite against Sonnet, but preserves the per-run ranking, so the composite's ranking is stable across judge vendors. Gemini-3-Pro was not included as a cross-judge at experiment time because of a quota block.

Throughout the paper, ``quality'' means the LLM-judge composite; its four dimensions do not measure maintainability, security, or long-term evolution cost, and the composite should not be read as software quality at large.

\subsection{Tasks}
\label{sec:tasks}

\textbf{Task suite.} For the main task suite we use four Express.js/TypeScript tasks that span a range of difficulty and distinct domain types; Appendix~\ref{app:tasks} gives the T1/T2/T4 specs. T1 asks for JWT authentication across six files. T2 is a marketplace API with $\geq$10 files. T4 demands a double-entry financial ledger with multi-currency, hash-chained audit~\citep{haber_timestamp}, reconciliation, and fraud detection, over $\geq$15 files. T5 pushes further, to an algorithmic trading platform with order matching, portfolio P\&L, risk, market data, and accounts, again over $\geq$15 files.
\textbf{Scope and budgets.}
We use a fifth task T3 (collaborative e-commerce modules) only in the difficulty-gradient appendix (Appendix~\ref{app:more_results}), and exclude it from the CMH pool.
The four tasks span different application domains at a larger scope each time, and they share an Express.js/TypeScript runtime and the same sandbox template, but not the application logic.
T1, T2 and T3 get 300\,s of wall clock, T4 gets 600\,s, and T5 gets 900\,s.
\textbf{Evaluation.}
The agents author their own vitest suites on top of the single smoke test in the template, so we report raw pass counts only descriptively. We assess correctness and quality with our LLM-judge composite, then cross-validate it against the regex and AST scorers (\S\ref{sec:scoring}).
We report a small Python cross-domain check using DevBench~\citep{devbench} (now DevEval) in Appendix~\ref{app:more_results}. We leave broader cross-language evaluation as a limitation (\S\ref{sec:limitations}).
We draw the protocol as one run's swim lane in Appendix~\ref{app:room-logs} (Figure~\ref{fig:dialogue}).

\section{Experiments}
\label{sec:filter}

\textbf{Models.} We use five frontier coding models from three providers: Anthropic's Claude Sonnet~4.6 and Haiku~4.5, OpenAI's GPT-5.4 and GPT-5.4-mini, and Google's Gemini~3 Flash. The headline cells use the four CLI-stable models. We exclude GPT-5.4-mini, whose vendor CLI crashes under concurrent MCP execution (\S\ref{sec:cli_fragility}); the crash is in the vendor's CLI, not in AgentRoom.
\textbf{Access and pinning.} We access the models through their vendor CLIs in non-interactive mode with vendor-default sampling, with versions pinned (Claude Code 2.1.119, Codex CLI 0.120.0, Gemini CLI 0.40.1; the per-run logs record model identifiers and per-agent invocation strings). We set no seeds: the models are proprietary and hosted, so we do not claim bit-for-bit reproducibility; we report run-to-run spread as $\sigma$ throughout.
This is a hosted-model ceiling on replication: hosted models can change server-side, but every contrast is matched-CLI and matched-budget, so drift moves both arms together.

\textbf{Conditions.} In \emph{AgentRoom}, our $N$ agents launch within 1\,s and share $\mathcal{W}$ and the room MCP server.
\emph{Shared-only} keeps the same launch but drops the collab prompt and the MCP.
\emph{Parallel-merge}: $N$ solo agents in separate workspaces, post-hoc file union with later-timestamp tie-break.

\textbf{Valid-run filter.} We keep all attempts that ran at least 30\,s. This excludes only the authentication and network failures that exit within the first 30\,s.
We apply no further filters on file count or test count: our earlier analyses found them biased against the compact implementations of Codex and Gemini.

Headline cells report $\sigma$ where $n \geq 3$. Our Tier II LLM-judge cells come from the budget-fair pool; a few runs that completed after we froze the pool point the same way, but we do not fold them into the reported cells.
Each table lists its own per-cell $n$, and the appendix flags its underpowered cells as exploratory.

\paragraph{Metrics.} For each run we measure the continuous quality score $s \in [0,1]$ produced by the LLM-judge of \S\ref{sec:scoring} applied to the post-run repository snapshot; the emergence score $\mathrm{ES}$ of \S\ref{sec:scoring}; raw passing-test count from vitest; room-message count (broadcast and direct-send entries); CRDT-conflict count (guard firings within $\Delta_{\text{crdt}}{=}2$\,s).

\paragraph{Evidence tiers.} We sort the results into three evidence tiers: Tier~I, the judge-free binary abandonment outcome; Tier~II, the continuous LLM-judge quality cells; and Tier~III, the exploratory comparisons across models under the regex/AST scorer family. Cells outside the T4 LLM-scored pool (cross-model effect size, difficulty gradient) are Tier~III.

\section{Results}
\label{sec:results}

We report the results in two tiers of decreasing statistical strength.
Tier I is the reduction in variance and abandonment rate, which replicates across all three powered CLI-stable models with a binary outcome that does not depend on the scorer. Tier II is the controlled six-condition ablation and the ChatDev paradigm contrast under our LLM-judge rubric (matched compute on the T4 fintech-ledger task).
The cross-model ES under the regex/AST scorer family and the cost-equivalent paired-model comparison are exploratory and reported in Appendix~\ref{app:more_results}.

\subsection{Tier I: Variance and Abandonment-Rate Reduction at $\times 2$}
\label{sec:variance}

\begin{figure}[t]
\centering
\includegraphics[width=\linewidth]{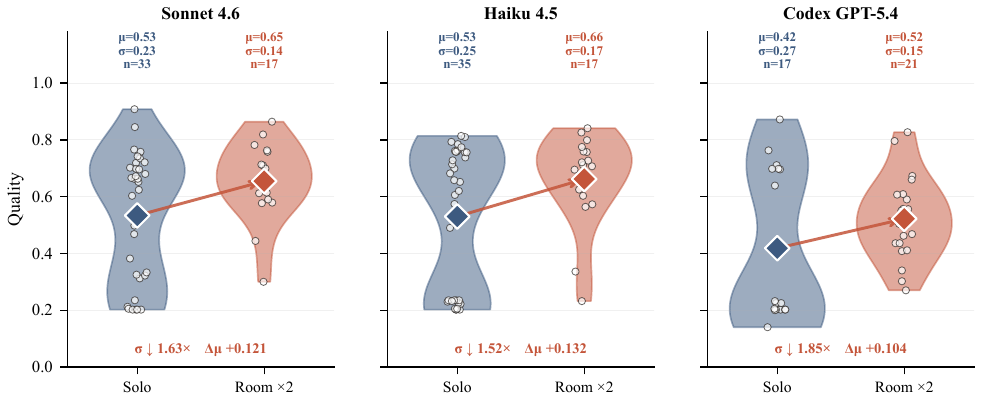}
\caption{Solo (slate) vs AgentRoom~$\times 2$ (rust) score distributions for the three powered CLI-stable models on T4 (LLM-judge composite, the budget-fair pool of the abandonment analysis in \S\ref{sec:variance}, which also holds the runs added after Table~\ref{tab:ablation}'s cells were frozen). $\sigma$ contracts by $\sim 30$--$45$\% under AgentRoom in every panel (Sonnet $0.23{\to}0.14$, Haiku $0.25{\to}0.17$, Codex $0.27{\to}0.15$); mean shift varies by model.}
\label{fig:variance}
\end{figure}

Two agents in an AgentRoom cut the run-to-run standard deviation by ${\sim}30$--$45\%$ among in-budget non-failing runs (Figure~\ref{fig:variance}) and suppress a specific lone-agent failure mode, \emph{1-file abandonment}: a run below $0.3$ that exits within $200$\,s or leaves at most two source files, typically a single stub emitted in the first ${\sim}80$\,s.
We see the same result on all three powered CLI-stable models (Sonnet 4.6, Haiku 4.5, Codex GPT-5.4) and on both Tier I tasks (T4, T5).
On T4 Sonnet, $\sigma$ falls from $0.23$ for solo to $0.14$ for $\times 2$, and on T5 Sonnet from $0.23$ to $0.13$.
The variance-reduction direction is consistent with self-consistency~\citep{self_consistency} and mixture-of-agents ensembling~\citep{mixture_of_agents}, but AgentRoom achieves it through coordinated concurrent generation in a shared workspace rather than independent sampling and aggregation.
$\times 2$ also lifts mean quality, but that lift is noisier, so we report it as exploratory in Appendix~\ref{app:more_results}.

\paragraph{Per-model and per-task abandonment.} A deterministic codebook, applied to the snapshot and duration of each run, labels each run whose regex-scorer composite is below $0.3$ as either $1$-file abandonment or infrastructure failure.
In Appendix~\ref{app:codebook_sensitivity} we describe the codebook and sweep its thresholds.
The counts below compare Solo against AgentRoom~$\times 2$ on the budget-fair pool under Fisher's exact one-sided test~\citep{fisher1922}.
We count every budget-fair run in this pool, including runs without an LLM score, so the denominators differ slightly from Table~\ref{tab:ablation}.
On T4 the reduction is significant on its own for two models: Haiku falls from $12/35$ to $1/17$ ($\mathbf{p{=}0.025}$), and Codex from $9/17$ to $2/21$ ($\mathbf{p{=}0.005}$).
The remaining two are directionally consistent.
Sonnet drops from $6/33$ to $0/17$ ($p{=}0.070$), and post-MCP-fix Gemini (\S\ref{sec:cli_fragility}) from $2/5$ to $0/5$ ($p{=}0.22$).

\paragraph{Across tasks.} On T5 all three per-model strata are directionally consistent but individually underpowered (per-stratum counts in Figure~\ref{fig:forest}).
Across all seven per-model strata every Solo abandonment count is $\geq$ the AgentRoom count; two of four are individually significant on T4.

\paragraph{Pooled across strata.} We stratify by both model and task across all four Express.js task difficulties (T1/T2/T4/T5), which gives 12 model$\times$task strata over the three powered models.
We apply the deterministic classifier to all budget-fair runs, and the Cochran-Mantel-Haenszel test~\citep{mantel_haenszel} gives a common odds ratio of $\mathbf{13.7}$ ($95\%$ CI $[3.9, 48]$), $\chi^2{=}22.8$, $\mathbf{p<10^{-5}}$, with Tarone homogeneity $p{=}0.92$~\citep{tarone}. The per-stratum effects are statistically indistinguishable.

\paragraph{Robustness of the pool.} The small strata do not carry the pooled effect: T4-only CMH (3 strata) gives OR$=11.9$ [$3.0$, $47$], $p{=}5 \times 10^{-5}$.
Pooled per task, abandonment falls in every task: T4 $32\%{\to}5\%$ ($p{=}0.0001$) and T5 $23\%{\to}0\%$ ($p{=}0.031$) are significant on their own, and T1 and T2 point the same way at small $n$ (Figure~\ref{fig:forest}).
The flat Fisher's exact pooled count across all 12 strata ($40/131$ Solo $= 31\%$ vs $5/94$ AgentRoom $= 5\%$, $p{<}10^{-6}$) is shown for completeness.

\paragraph{Conditional quality.} On T4 Sonnet under the LLM-judge rubric, and excluding runs with catastrophic infrastructure failures, Solo averages $0.544$ ($n{=}32$) and AgentRoom $0.669$ ($n{=}14$), a lead of $+0.125$; Welch's test~\citep{welch} gives $t{=}2.30$, $p{=}0.027$.
AgentRoom's catastrophic-failure rate on Sonnet T4 matches Solo's because both encounter identical infrastructure failure rates; AgentRoom suppresses the abandonment failure mode (1-file stub-and-exit).

\subsection{Tier II.a: Six-Condition Ablation}
\label{sec:ablation}

\paragraph{Matched-compute ordering.} Table~\ref{tab:ablation} (T4, Sonnet 4.6, budget-fair pool $30$--$700$\,s) compares AgentRoom against five alternatives under the LLM-judge rubric: solo at the same wall-clock budget as the pairs, and the other four at matched compute.
The six conditions form a clean monotonic ordering---ChatDev-style sequential pipeline $0.333$ ($n{=}6$) $\ll$ parallel-merge $0.456$ ($n{=}12$) $<$ solo $0.544$ ($n{=}32$) $<$ shared-only $0.575$ ($n{=}11$) $<$ shared+collab without MCP $0.588$ ($n{=}7$) $<$ AgentRoom $0.669$ ($n{=}14$, $\sigma{=}0.140$).

\paragraph{Parallel-merge contrast.} The most informative single contrast is parallel-merge (concurrent without coordination) vs AgentRoom (concurrent with CRDT+MCP coordination); both run two Sonnet agents in parallel for the same wall-clock budget, but only AgentRoom has the shared CRDT workspace and explicit MCP signalling.
AgentRoom leads parallel-merge by $+0.213$ in mean quality, Welch $t{=}3.35$, $\mathbf{p{=}0.003}$.
Naive concurrent ensembling actually \emph{underperforms} a single agent ($0.456$ vs solo $0.544$) because without coordination the second agent's \texttt{server.ts} silently overwrites the first agent's file; parallel-merge amplifies rather than averages the stub-and-exit behavior.
The shared-only condition ($0.575$) sits between solo and AgentRoom, recovering some of the loss but not reaching the full bundle.

\paragraph{Cross-model check.} Appendix~\ref{app:more_results}, Table~\ref{tab:ablation_xmodel}, reports the cross-model ablation on Haiku and Codex: Haiku replicates the pattern with AgentRoom $0.662$ ($n{=}17$) above solo $0.530$ ($n{=}35$); Codex has the lowest solo mean there at $0.384$, and AgentRoom lifts it to $0.529$ ($n{=}17$); on T5 its pairs regress (Appendix~\ref{app:t5_matrix}), consistent with its monolithic-architecture priors resisting a split across two agents.

\begin{table}[h]
\centering
\caption{T4 Sonnet 4.6 six-condition ablation (LLM-judge composite, budget-fair $30$--$700$\,s pool). Parallel-merge vs AgentRoom: $+0.213$, Welch $t{=}3.35$, $\mathbf{p{=}0.003}$. The shared+collab-noMCP probe ran under a $1200$\,s wall budget; its runs are filtered to the same $30$--$700$\,s envelope as every other cell (Appendix~\ref{app:bundle_probe}). Cross-model replication in Appendix~\ref{app:more_results}, Table~\ref{tab:ablation_xmodel}.}
\label{tab:ablation}
\small
\begin{tabular}{lcccl}
\toprule
Condition & Mean & $n$ & $\sigma$ & Note \\
\midrule
ChatDev-style (sequential 3-phase) & 0.333 & 6 & 0.197 & sequential paradigm baseline \\
Parallel merge (2 indep + union) & 0.456 & 12 & 0.178 & concurrent, no coordination \\
Solo (1 agent) & 0.544 & 32 & 0.222 & single-agent baseline \\
Shared only (2 agents, no MCP) & 0.575 & 11 & 0.155 & shared CRDT, no signalling \\
Shared + collab prompt, noMCP & 0.588 & 7 & 0.188 & + prompt, no MCP tools \\
\midrule
\textbf{AgentRoom (2 agents, full)} & \textbf{0.669} & \textbf{14} & \textbf{0.140} & CRDT + room MCP + collab prompt \\
\bottomrule
\end{tabular}
\end{table}

\subsection{Tier II.b: Cross-Paradigm Comparison vs Sequential Role Pipeline}
\label{sec:chatdev}

Most multi-agent coding systems~\citep{chatdev,metagpt,agilecoder} use sequential turn-taking, each agent waiting for the previous. To compare against that paradigm, we built our own ChatDev-style baseline on \emph{the same model and CLI} as AgentRoom (Sonnet 4.6 via Claude Code) and did not import the published 2023 ChatDev framework, whose client supports only OpenAI and whose dependencies are no longer maintained, which would have introduced a model confound and version drift.
The baseline implements ChatDev's three-phase workflow faithfully (requirements analysis $\to$ technical design $\to$ implementation, with role-specific prompts and intermediate artifacts handed across phases) at a $1200$\,s budget, which is the compute of AgentRoom~$\times 2$.
Of the $13$ budget-fair ChatDev runs, $7$ are infrastructure failures: every phase agent exits with code $1$ and no output at ${\sim}32$\,s because the orchestrator crashed before any code was produced. We exclude these orchestrator crashes from the quality comparison, and no run in any other condition crashes this way.
On the six genuine ChatDev runs that remain (Sonnet, T4, budget-fair), mean quality rises from $\bar s{=}0.333$ for ChatDev ($\sigma{=}0.197$) to $0.669$ for AgentRoom~$\times 2$ ($n{=}14$, $\sigma{=}0.140$) on the same task at the same compute (Welch $t{=}3.78$, two-sided $p{=}0.006$).
In our six genuine ChatDev runs, $4$ abandoned the task in the implementation phase with a one-file source skeleton that still passes the boilerplate \texttt{GET /} test, while none of the $10$ AgentRoom~$\times 2$ runs we inspected by hand did; sequential pipelines amplify the lone-agent ``stub-and-exit'' failure mode at every phase boundary.
The ChatDev contrast is further evidence that AgentRoom's coordination layer suppresses this failure mode however the upstream task is structured; it is not evidence that ChatDev pipelines fail catastrophically more often overall.

\subsection{Tier II.c: Concurrent-MCP CLI Compatibility (deployment caveat)}
\label{sec:cli_fragility}

\textbf{Gemini.} Gemini~3 and Codex-mini both crashed when MCP ran concurrently. We traced Gemini's failure to a protocol non-conformance: our room MCP server emitted LSP-style \texttt{Content-Length:} framing while Gemini~CLI~0.40+ strictly enforces the MCP-stdio NDJSON spec; switching the room server to NDJSON (a one-line server-side change) recovered Gemini in full. A post-fix Gemini~T4 batch of five solo runs and five Gemini pairs reproduces our headline pattern---Solo abandonment $2/5{=}40\%$ (1-file stub-and-exit at 30--50\,s) collapses to $0/5$ under AgentRoom, with mean LLM-judge composite $0.519$ ($\sigma{=}0.266$) $\to$ $0.673$ ($\sigma{=}0.056$) and 5/5 runs passing vitest.
\textbf{Codex-mini.} Unlike Gemini, Codex-mini still crashes under concurrent MCP execution, so we exclude it from all cells.

\subsection{Scaling and Heterogeneity}
\label{sec:scaling}

\begin{figure}[t]
\centering
\includegraphics[width=\linewidth]{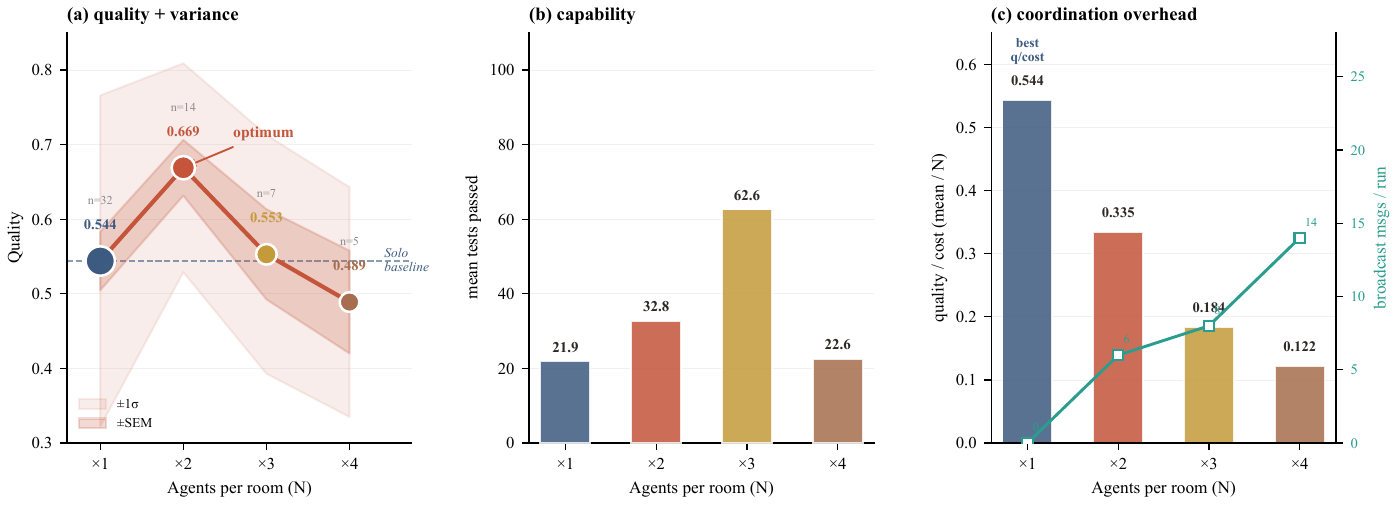}
\caption{Agent-count scaling on T4 Sonnet 4.6 ($N{=}1$--$4$, budget-fair pool). Left: mean LLM-judge composite with $\sigma$ band peaks at $N{=}2$ then declines. Middle: mean tests passing peaks at $N{=}3$. Right: AgentRoom message count rises super-linearly past $N{=}3$, consistent with a single-broadcast-channel coordination overhead.}
\label{fig:scaling}
\end{figure}

Sonnet T4 quality peaks at $\times 2$ then declines under our LLM-judge rubric---$0.544$ / $0.669$ / $0.553$ / $0.489$ for $\times 1/\times 2/\times 3/\times 4$ on the budget-fair pool---while mean tests passing peaks at $\times 3$ (Figure~\ref{fig:scaling}, middle panel).
Per-run wall-clock stays near the task budget across $N$ (Appendix~\ref{app:compute}), so total compute grows roughly linearly with $N$. The message count rises super-linearly past $\times 3$ (Figure~\ref{fig:scaling}, right panel). We have not run a controlled multi-channel ablation, so whether the decline reflects saturation of the single broadcast channel is an open question rather than an attribution, and the separating ablation is left for future work. Two agents work best in our setup; $\times 3$ and $\times 4$ pay linearly more compute for a decline in mean quality.
\textbf{Heterogeneous pairs.} On T4, heterogeneous pairing (Sonnet primary partner) puts $\times 2$ Sonnet+Codex at $0.721$ ($n{=}3$), the highest of the heterogeneous pairings we tested, while $\times 2$ Sonnet+Gemini scores only $0.542$ ($n{=}3$). Even with the first agent held fixed, one second agent contributes more than the other, which suggests that mixing models is not a free lunch and that the coordination layer does not remove differences between models (Table~\ref{tab:heterogeneous}).

\subsection{MCP Chat vs CRDT State}
\label{sec:chat_vs_state}

Comparing the room backends on T5 shows that our MCP-chat baseline ($n{=}2$) fails 1 of 2 runs, scoring 0.088 and passing zero tests, when the two agents broadcast their file claims simultaneously as free text and neither notices the conflict, while the CRDT-state AgentRoom ($n{=}7$) succeeds in 7 of its 7 runs because \texttt{room\_claim()} returns a system-level conflict error.
Our sample is small but our qualitative finding (one observed catastrophic chat failure, zero CRDT-state failures) motivates a larger study.

\section{Discussion}
\label{sec:discussion}

\textbf{Coordination, not concurrency, drives the AgentRoom gain.}
The full six-condition ordering on T4 Sonnet (Table~\ref{tab:ablation}) and the parallel-merge vs AgentRoom contrast ($+0.213$, Welch $t{=}3.35$, $p{=}0.003$) are what \S\ref{sec:ablation} reports.
The bundle is co-designed, and a probe separates its components only partially: keeping the CRDT substrate and collaboration prompt but removing the MCP tool surface (Appendix~\ref{app:bundle_probe}, $n{=}7$) lands at $0.588$, $+0.013$ above shared-only. The MCP-tool step accounts for the remaining $+0.081$ up to AgentRoom's $0.669$. At these cell sizes the interval on that step spans zero, so we claim the observed mean ordering shared-only $<$ prompt-only $<$ AgentRoom rather than a percentage split. The ordering still argues against reading AgentRoom as ``CRDT plus prompts'' integration: the largest observed step is the explicit coordination channel; the substrate is the merge primitive on which it stands.
AgentRoom's claim protocol creates an advisory distributed lock by prompt-level cooperation; the advisory-plus-apology pattern matters because strict locking would block the cross-agent bug-fix case (a teammate correcting an Express~5 routing error in another agent's file, Appendix~\ref{app:room-logs}).
The design lineage is advisory locking in distributed systems~\citep{chubby,zookeeper,leases}; AgentRoom's file-level claim is a Chubby-style advisory lock surfaced as an MCP tool, with the agent rather than a kernel call as the cooperator. We deliberately avoid consensus protocols~\citep{raft} because CRDT convergence already provides the safety property advisory claims need; causality is established by the underlying CRDT~\citep{shapiro_crdt} which inherits Lamport's happens-before~\citep{lamport_clocks} for ordering concurrent edits, in contrast to the operational-transformation lineage~\citep{ellis_gibbs_ot,sun_cci}, which transforms concurrent operations against each other.

\textbf{Cost-equivalent comparison is suggestive.}
$2\times$Haiku AgentRoom reaches T4 mean $0.662$ ($n{=}17$) versus $1\times$Sonnet $0.544$ ($n{=}32$) at about two-thirds of the dollar cost---Welch $t{=}2.10$, $p{=}0.041$. This is an actionable hint that paired smaller models can be Pareto-competitive in this difficulty band: the full cross-model headroom matrix (5/6 cells point-estimate-positive, only Haiku T4 clears its bootstrap CI) is what Appendix~\ref{app:more_results} reports.

\textbf{Activation of collaborative language.}
AgentRoom logs contain unprompted patterns---apologies for boundary violations, proactive gap-filling, ``do not'' warnings, dependency documentation (Appendix~\ref{app:more_results}, Appendix~\ref{app:room-logs}). We replay the archived room logs pool-wide to quantify claim-discipline adherence against quality (Appendix~\ref{app:claim_coverage}).
Agents claim modules in 6 of 6 runs---they fix each other's bugs in only 2 of 6, and we do not treat that as the typical case.
The shared-only condition, which has neither the channel nor the collaboration prompt, produces none of these; we describe the patterns and do not attempt to attribute them to a specific training-data source.

\textbf{Structured state vs unstructured chat.}
Comparing the room backends on T5 shows that MCP-chat ($n{=}2$) fails 1 of 2 runs (score 0.088) when two agents broadcast file claims as free text and neither parses the conflict; CRDT-state AgentRoom ($n{=}7$) succeeds 7/7 because \texttt{room\_claim()} returns a system-level conflict error. The state-management level enforces ownership deterministically where text-parsed claims do not.

\textbf{Paradigm contrast: AgentRoom vs ChatDev-style sequential.}
The full numbers and Welch test on the same-model same-CLI ChatDev-style three-phase pipeline at matched compute are what \S\ref{sec:chatdev} reports. The contrast isolates the paradigm choice between sequential phase-handoff and concurrent CRDT-coordinated workspace at the same model, CLI, task, and compute, and sequential pipelines amplify the lone-agent failure mode at every phase boundary.
A complementary prompt-vs-channel control on Haiku (Appendix~\ref{app:more_results}) is less conclusive because the structured tool prompt also acts as workflow scaffolding for Haiku and is not separately identified from channel use at this $n$.

\textbf{Cross-language and cross-domain replication.} Two out-of-distribution checks reproduce the direction beyond Express.js/TypeScript: on a Rust+axum transposition of T4 (Codex GPT-5.4) AgentRoom~$\times 2$ matches Solo on the LLM-judge composite ($0.740$ vs $0.714$, $n{=}4$ vs $5$, descriptive; Appendix~\ref{app:rust}), and on 10 DevBench Python multi-file projects (Haiku 4.5) $\times 2$ passes $8/10$ against Solo's $7/10$ with zero CRDT semantic conflicts (Appendix~\ref{app:devbench}). Each is individually underpowered---jointly the direction is not TypeScript-specific.

\textbf{Comparison with prior multi-agent coding systems.}
CodeCRDT~\citep{codecrdt} and AgentRoom share the CRDT-merged substrate but propose different coordination paradigms.
CodeCRDT is implicit and observation-driven: agents infer teammate state by reading the merged workspace, roles are pre-assigned (outliner / implementer); the empirical question is speedup over sequential baselines (reported bimodal at $21\%$ speedup / $39\%$ slowdown, semantic conflict rate $5$--$10\%$).
AgentRoom is explicit: agents negotiate ownership and intent through the room (file-level claim, broadcast log, per-agent status) via MCP tools rather than by re-reading file deltas; the empirical question is failure-mode suppression rather than speedup.
The two designs produce different signatures: CodeCRDT's character-level concurrent edits produce the $5$--$10\%$ semantic conflicts and the bimodal speed distribution, where AgentRoom's advisory file-level claim leaves $0\%$ semantic conflicts in the headline pool and the parallel-merge vs AgentRoom matched-compute contrast attributes the gain to the explicit signalling channel rather than to the CRDT substrate.
Concurrent industry deployments such as Cursor's multi-agent editor, Anthropic's Claude Code agent teams, and JetBrains AI Assistant Multi-Agent run the file-claim pattern in production; what we add here is the first matched-compute quantification isolating the coordination layer from parallelism and from the CRDT substrate, on a failure mode that neither CodeCRDT nor the shipping systems measure.
CodeCRDT and AgentRoom are complementary: CodeCRDT's character-level merge could power AgentRoom's intra-file editing within a claimed file. Other recent multi-agent coding systems~\citep{codecor,codesim,semag} share the LLM-team-coding goal but use sequential or staged-role paradigms.

\section{Conclusion}

Prior work has treated concurrent multi-agent coding as a parallelism problem or a merge-correctness problem; we hold that it is primarily about explicit coordination. In our matched-compute setting, neither agent count nor the CRDT itself is the main contributor we identify; the gain comes from the runtime room's state-management operations: claiming files, announcing progress, and querying the room state. Without coordination, naive concurrency underperforms a single agent; with the substrate alone, agents converge on bytes but not on intent. Within our setting, the substrate--coordination separation suggests that an explicit room-style primitive is the contributor we identify, and that matched-compute ablation is how we measure such contributions. We leave open whether this result generalizes beyond coding tasks.

\section{Limitations}
\label{sec:limitations}

The headline results span four backend domains and the three powered CLI-stable models from two providers, and we exclude Codex-mini (\S\ref{sec:cli_fragility}).
We replay a post-hoc capability probe over the archived T4 snapshots, and it corroborates the ends of the ordering (Appendix~\ref{app:oracle}).
All headline cells share one Express.js/TypeScript runtime.
The Rust and Python checks are bounded spot checks of mechanism portability rather than of quality magnitude.
In the parallel-merge vs AgentRoom contrast, Sonnet produces the code on both sides, and we cross-validate the LLM-judge composite against independent regex and TypeScript-AST scorers.
Our bundle-component probe reuses the ablation cells, so its substrate-vs-coordination split is only a rough guide to the variance.
Our comparison of two Haiku agents with one Sonnet agent is exposed to the Sonnet judge's same-family preference (\S\ref{sec:scoring}), so it is only suggestive.
We do not evaluate software in the wild, where no oracle exists.
Headline cells have $6$ to $35$ runs each; Gemini~3 is underpowered at $5{+}5$.
Correctness is checked by held-out tests only in the post-hoc probe; otherwise quality rests on the LLM-judge composite, since the agents write their own vitest suites.

\begin{ack}
This work received no external funding. The authors declare no competing interests.
\end{ack}

\bibliographystyle{plainnat}
\bibliography{references}

\begin{thebibliography}{41}
\providecommand{\natexlab}[1]{#1}
\providecommand{\url}[1]{\texttt{#1}}
\expandafter\ifx\csname urlstyle\endcsname\relax
  \providecommand{\doi}[1]{doi: #1}\else
  \providecommand{\doi}{doi: \begingroup \urlstyle{rm}\Url}\fi

\bibitem[Ashrafi et~al.(2025)Ashrafi, Bouktif, and
  Mediani]{multiagent_runtime_debug}
Nazmus Ashrafi, Salah Bouktif, and Mohammed Mediani.
\newblock Enhancing llm code generation: A systematic evaluation of multi-agent
  collaboration and runtime debugging for improved accuracy, reliability, and
  latency, 2025.
\newblock URL \url{https://arxiv.org/abs/2505.02133}.

\bibitem[{Automerge Project}(2023)]{automerge}
{Automerge Project}.
\newblock Automerge: A library of data structures for building collaborative
  applications, 2023.
\newblock URL \url{https://automerge.org}.

\bibitem[Burrows(2006)]{chubby}
Michael Burrows.
\newblock The chubby lock service for loosely-coupled distributed systems.
\newblock In Brian~N. Bershad and Jeffrey~C. Mogul, editors, \emph{7th
  Symposium on Operating Systems Design and Implementation {(OSDI} '06),
  November 6-8, Seattle, WA, {USA}}, pages 335--350. {USENIX} Association,
  2006.
\newblock URL \url{http://www.usenix.org/events/osdi06/tech/burrows.html}.

\bibitem[Chen et~al.(2023)Chen, Su, Zuo, Yang, Yuan, Chan, Yu, Lu, Hung, Qian,
  Qin, Cong, Xie, Liu, Sun, and Zhou]{agentverse}
Weize Chen, Yusheng Su, Jingwei Zuo, Cheng Yang, Chenfei Yuan, Chi-Min Chan,
  Heyang Yu, Yaxi Lu, Yi-Hsin Hung, Chen Qian, Yujia Qin, Xin Cong, Ruobing
  Xie, Zhiyuan Liu, Maosong Sun, and Jie Zhou.
\newblock Agentverse: Facilitating multi-agent collaboration and exploring
  emergent behaviors, 2023.
\newblock URL \url{https://arxiv.org/abs/2308.10848}.

\bibitem[Efron(1979)]{efron1979}
Bradley Efron.
\newblock Bootstrap methods: Another look at the jackknife.
\newblock \emph{The Annals of Statistics}, 7\penalty0 (1):\penalty0 1--26,
  1979.
\newblock \doi{10.1214/aos/1176344552}.

\bibitem[Ellis and Gibbs(1989)]{ellis_gibbs_ot}
C.~A. Ellis and S.~J. Gibbs.
\newblock Concurrency control in groupware systems.
\newblock In \emph{Proceedings of the 1989 ACM SIGMOD International Conference
  on Management of Data}, SIGMOD '89, pages 399--407, New York, NY, USA, 1989.
  Association for Computing Machinery.
\newblock ISBN 0897913175.
\newblock \doi{10.1145/67544.66963}.
\newblock URL \url{https://doi.org/10.1145/67544.66963}.

\bibitem[Fisher(1922)]{fisher1922}
Ronald~A. Fisher.
\newblock On the interpretation of $\chi^2$ from contingency tables, and the
  calculation of {P}.
\newblock \emph{Journal of the Royal Statistical Society}, 85\penalty0
  (1):\penalty0 87--94, 1922.
\newblock \doi{10.2307/2340521}.

\bibitem[Gray and Cheriton(1989)]{leases}
Cary~G. Gray and David~R. Cheriton.
\newblock Leases: An efficient fault-tolerant mechanism for distributed file
  cache consistency.
\newblock In Gregory~R. Andrews, editor, \emph{Proceedings of the Twelfth {ACM}
  Symposium on Operating System Principles, {SOSP} 1989, The Wigwam, Litchfield
  Park, Arizona, USA, December 3-6, 1989}, pages 202--210. {ACM}, 1989.
\newblock \doi{10.1145/74850.74870}.
\newblock URL \url{https://doi.org/10.1145/74850.74870}.

\bibitem[Grötschla et~al.(2025)Grötschla, Müller, Tönshoff, Galkin, and
  Perozzi]{agentsnet}
Florian Grötschla, Luis Müller, Jan Tönshoff, Mikhail Galkin, and Bryan
  Perozzi.
\newblock Agentsnet: Coordination and collaborative reasoning in multi-agent
  llms, 2025.
\newblock URL \url{https://arxiv.org/abs/2507.08616}.

\bibitem[Haber and Stornetta(1991)]{haber_timestamp}
Stuart Haber and W.~Scott Stornetta.
\newblock How to time-stamp a digital document.
\newblock \emph{Journal of Cryptology}, 3\penalty0 (2):\penalty0 99--111, 1991.
\newblock \doi{10.1007/BF00196791}.
\newblock URL \url{https://doi.org/10.1007/BF00196791}.

\bibitem[Hong et~al.(2024)Hong, Zhuge, Chen, Zheng, Cheng, Zhang, Wang, Wang,
  Yau, Lin, Zhou, Ran, Xiao, Wu, and Schmidhuber]{metagpt}
Sirui Hong, Mingchen Zhuge, Jiaqi Chen, Xiawu Zheng, Yuheng Cheng, Ceyao Zhang,
  Jinlin Wang, Zili Wang, Steven Ka~Shing Yau, Zijuan Lin, Liyang Zhou, Chenyu
  Ran, Lingfeng Xiao, Chenglin Wu, and Jürgen Schmidhuber.
\newblock Metagpt: Meta programming for a multi-agent collaborative framework,
  2024.
\newblock URL \url{https://arxiv.org/abs/2308.00352}.

\bibitem[Huang et~al.(2024)Huang, Zhang, Luck, Bu, Qing, and Cui]{agentcoder}
Dong Huang, Jie~M. Zhang, Michael Luck, Qingwen Bu, Yuhao Qing, and Heming Cui.
\newblock Agentcoder: Multi-agent-based code generation with iterative testing
  and optimisation, 2024.
\newblock URL \url{https://arxiv.org/abs/2312.13010}.

\bibitem[Hunt et~al.(2010)Hunt, Konar, Junqueira, and Reed]{zookeeper}
Patrick Hunt, Mahadev Konar, Flavio~P. Junqueira, and Benjamin Reed.
\newblock {ZooKeeper}: Wait-free coordination for internet-scale systems.
\newblock In \emph{2010 USENIX Annual Technical Conference (USENIX ATC 10)}.
  USENIX Association, June 2010.
\newblock URL
  \url{https://www.usenix.org/conference/usenix-atc-10/zookeeper-wait-free-coordination-internet-scale-systems}.

\bibitem[Ishibashi and Nishimura(2024)]{selforganized}
Yoichi Ishibashi and Yoshimasa Nishimura.
\newblock Self-organized agents: A llm multi-agent framework toward ultra
  large-scale code generation and optimization, 2024.
\newblock URL \url{https://arxiv.org/abs/2404.02183}.

\bibitem[Islam et~al.(2025)Islam, Ali, and Parvez]{codesim}
Md.~Ashraful Islam, Mohammed~Eunus Ali, and Md~Rizwan Parvez.
\newblock Codesim: Multi-agent code generation and problem solving through
  simulation-driven planning and debugging, 2025.
\newblock URL \url{https://arxiv.org/abs/2502.05664}.

\bibitem[Jahns(2024)]{yjs}
Kevin Jahns.
\newblock {Yjs}: A {CRDT} framework for shared editing.
\newblock \url{https://github.com/yjs/yjs}, 2024.

\bibitem[Kleppmann and Beresford(2017)]{json_crdt}
Martin Kleppmann and Alastair~R. Beresford.
\newblock A conflict-free replicated json datatype, 2017.
\newblock URL \url{https://arxiv.org/abs/1608.03960}.

\bibitem[Lamport(1978)]{lamport_clocks}
Leslie Lamport.
\newblock Time, clocks, and the ordering of events in a distributed system.
\newblock \emph{Communications of the ACM}, 21\penalty0 (7):\penalty0 558--565,
  July 1978.
\newblock ISSN 1557-7317.
\newblock \doi{10.1145/359545.359563}.
\newblock URL \url{http://dx.doi.org/10.1145/359545.359563}.

\bibitem[Li et~al.(2024)Li, Wu, Tang, Shi, Yang, Li, Yao, Qian, Hui, Zhang, Yu,
  Du, Yang, Lin, Peng, and Chen]{devbench}
Bowen Li, Wenhan Wu, Ziwei Tang, Lin Shi, John Yang, Jinyang Li, Shunyu Yao,
  Chen Qian, Binyuan Hui, Qicheng Zhang, Zhiyin Yu, He~Du, Ping Yang, Dahua
  Lin, Chao Peng, and Kai Chen.
\newblock Prompting large language models to tackle the full software
  development lifecycle: A case study, 2024.
\newblock URL \url{https://arxiv.org/abs/2403.08604}.

\bibitem[Li et~al.(2026)Li, Sun, Huang, Zhong, Jiang, Han, Zhang, Wang, and
  Liu]{preference_leakage}
Dawei Li, Renliang Sun, Yue Huang, Ming Zhong, Bohan Jiang, Jiawei Han,
  Xiangliang Zhang, Wei Wang, and Huan Liu.
\newblock Preference leakage: A contamination problem in llm-as-a-judge, 2026.
\newblock URL \url{https://arxiv.org/abs/2502.01534}.

\bibitem[Mantel and Haenszel(1959)]{mantel_haenszel}
Nathan Mantel and William Haenszel.
\newblock Statistical aspects of the analysis of data from retrospective
  studies of disease.
\newblock \emph{Journal of the National Cancer Institute}, 22\penalty0
  (4):\penalty0 719--748, 1959.
\newblock \doi{10.1093/jnci/22.4.719}.

\bibitem[Nguyen et~al.(2024)Nguyen, Chau, Nguyen, and Bui]{agilecoder}
Minh~Huynh Nguyen, Thang~Phan Chau, Phong~X. Nguyen, and Nghi D.~Q. Bui.
\newblock Agilecoder: Dynamic collaborative agents for software development
  based on agile methodology, 2024.
\newblock URL \url{https://arxiv.org/abs/2406.11912}.

\bibitem[Ongaro and Ousterhout(2014)]{raft}
Diego Ongaro and John~K. Ousterhout.
\newblock In search of an understandable consensus algorithm.
\newblock In Garth Gibson and Nickolai Zeldovich, editors, \emph{Proceedings of
  the 2014 {USENIX} Annual Technical Conference, {USENIX} {ATC} 2014,
  Philadelphia, PA, USA, June 19-20, 2014}, pages 305--319. {USENIX}
  Association, 2014.
\newblock URL
  \url{https://www.usenix.org/conference/atc14/technical-sessions/presentation/ongaro}.

\bibitem[Pan et~al.(2026)Pan, Zhang, and Liu]{codecor}
Ruwei Pan, Hongyu Zhang, and Chao Liu.
\newblock Codecor: An llm-based self-reflective multi-agent framework for code
  generation, 2026.
\newblock URL \url{https://arxiv.org/abs/2501.07811}.

\bibitem[Park et~al.(2023)Park, O'Brien, Cai, Morris, Liang, and
  Bernstein]{generative_agents}
Joon~Sung Park, Joseph~C. O'Brien, Carrie~J. Cai, Meredith~Ringel Morris, Percy
  Liang, and Michael~S. Bernstein.
\newblock Generative agents: Interactive simulacra of human behavior, 2023.
\newblock URL \url{https://arxiv.org/abs/2304.03442}.

\bibitem[Peng et~al.(2026)Peng, Hou, Zhu, He, and Yu]{semag}
Yulin Peng, Haowen Hou, Xinxin Zhu, Ying~Tiffany He, and F.~Richard Yu.
\newblock Semag: Self-evolutionary multi-agent code generation, 2026.
\newblock URL \url{https://arxiv.org/abs/2603.15707}.

\bibitem[Pugachev(2025)]{codecrdt}
Sergey Pugachev.
\newblock Codecrdt: Observation-driven coordination for multi-agent llm code
  generation, 2025.
\newblock URL \url{https://arxiv.org/abs/2510.18893}.

\bibitem[Qian et~al.(2024)Qian, Liu, Liu, Chen, Dang, Li, Yang, Chen, Su, Cong,
  Xu, Li, Liu, and Sun]{chatdev}
Chen Qian, Wei Liu, Hongzhang Liu, Nuo Chen, Yufan Dang, Jiahao Li, Cheng Yang,
  Weize Chen, Yusheng Su, Xin Cong, Juyuan Xu, Dahai Li, Zhiyuan Liu, and
  Maosong Sun.
\newblock Chatdev: Communicative agents for software development, 2024.
\newblock URL \url{https://arxiv.org/abs/2307.07924}.

\bibitem[Roh et~al.(2011)Roh, Jeon, Kim, and Lee]{rga}
Hyun-Gul Roh, Myeongjae Jeon, Jin-Soo Kim, and Joonwon Lee.
\newblock Replicated abstract data types: Building blocks for collaborative
  applications.
\newblock \emph{Journal of Parallel and Distributed Computing}, 71\penalty0
  (3):\penalty0 354--368, 2011.
\newblock ISSN 0743-7315.
\newblock \doi{10.1016/j.jpdc.2010.12.006}.
\newblock URL
  \url{https://www.sciencedirect.com/science/article/pii/S0743731510002716}.

\bibitem[Shafin et~al.(2025)Shafin, Rafi, Li, and
  Chen]{process_models_multiagent}
Wasique~Islam Shafin, Md~Nakhla Rafi, Zhenhao Li, and Tse-Hsun Chen.
\newblock Evaluating software process models for multi-agent class-level code
  generation, 2025.
\newblock URL \url{https://arxiv.org/abs/2511.09794}.

\bibitem[Shapiro et~al.(2011)Shapiro, Preguiça, Baquero, and
  Zawirski]{shapiro_crdt}
Marc Shapiro, Nuno Preguiça, Carlos Baquero, and Marek Zawirski.
\newblock Conflict-free replicated data types.
\newblock In \emph{Stabilization, Safety, and Security of Distributed Systems},
  pages 386--400. Springer Berlin Heidelberg, 2011.
\newblock ISBN 9783642245503.
\newblock \doi{10.1007/978-3-642-24550-3_29}.
\newblock URL \url{http://dx.doi.org/10.1007/978-3-642-24550-3_29}.

\bibitem[Shinn et~al.(2023)Shinn, Cassano, Berman, Gopinath, Narasimhan, and
  Yao]{reflexion}
Noah Shinn, Federico Cassano, Edward Berman, Ashwin Gopinath, Karthik
  Narasimhan, and Shunyu Yao.
\newblock Reflexion: Language agents with verbal reinforcement learning, 2023.
\newblock URL \url{https://arxiv.org/abs/2303.11366}.

\bibitem[Sun et~al.(1998)Sun, Jia, Zhang, Yang, and Chen]{sun_cci}
Chengzheng Sun, Xiaohua Jia, Yanchun Zhang, Yun Yang, and David Chen.
\newblock Achieving convergence, causality preservation, and intention
  preservation in real-time cooperative editing systems.
\newblock \emph{ACM Trans. Comput.-Hum. Interact.}, 5\penalty0 (1):\penalty0
  63--108, March 1998.
\newblock ISSN 1073-0516.
\newblock \doi{10.1145/274444.274447}.
\newblock URL \url{https://doi.org/10.1145/274444.274447}.

\bibitem[Tarone(1985)]{tarone}
Robert~E. Tarone.
\newblock On heterogeneity tests based on efficient scores.
\newblock \emph{Biometrika}, 72\penalty0 (1):\penalty0 91--95, 1985.
\newblock \doi{10.1093/biomet/72.1.91}.

\bibitem[Wang et~al.(2024)Wang, Wang, Athiwaratkun, Zhang, and
  Zou]{mixture_of_agents}
Junlin Wang, Jue Wang, Ben Athiwaratkun, Ce~Zhang, and James Zou.
\newblock Mixture-of-agents enhances large language model capabilities, 2024.
\newblock URL \url{https://arxiv.org/abs/2406.04692}.

\bibitem[Wang et~al.(2025)Wang, Li, Song, Xu, Tang, Zhuge, Pan, Song, Li,
  Singh, Tran, Li, Ma, Zheng, Qian, Shao, Muennighoff, Zhang, Hui, Lin,
  Brennan, Peng, Ji, and Neubig]{openhands}
Xingyao Wang, Boxuan Li, Yufan Song, Frank~F. Xu, Xiangru Tang, Mingchen Zhuge,
  Jiayi Pan, Yueqi Song, Bowen Li, Jaskirat Singh, Hoang~H. Tran, Fuqiang Li,
  Ren Ma, Mingzhang Zheng, Bill Qian, Yanjun Shao, Niklas Muennighoff, Yizhe
  Zhang, Binyuan Hui, Junyang Lin, Robert Brennan, Hao Peng, Heng Ji, and
  Graham Neubig.
\newblock Openhands: An open platform for ai software developers as generalist
  agents, 2025.
\newblock URL \url{https://arxiv.org/abs/2407.16741}.

\bibitem[Wang et~al.(2023)Wang, Wei, Schuurmans, Le, Chi, Narang, Chowdhery,
  and Zhou]{self_consistency}
Xuezhi Wang, Jason Wei, Dale Schuurmans, Quoc Le, Ed~Chi, Sharan Narang,
  Aakanksha Chowdhery, and Denny Zhou.
\newblock Self-consistency improves chain of thought reasoning in language
  models, 2023.
\newblock URL \url{https://arxiv.org/abs/2203.11171}.

\bibitem[Wataoka et~al.(2025)Wataoka, Takahashi, and Ri]{self_preference}
Koki Wataoka, Tsubasa Takahashi, and Ryokan Ri.
\newblock Self-preference bias in llm-as-a-judge, 2025.
\newblock URL \url{https://arxiv.org/abs/2410.21819}.

\bibitem[Welch(1947)]{welch}
B.~L. Welch.
\newblock The generalization of {`Student's'} problem when several different
  population variances are involved.
\newblock \emph{Biometrika}, 34\penalty0 (1-2):\penalty0 28--35, 1947.
\newblock \doi{10.1093/biomet/34.1-2.28}.

\bibitem[Wu et~al.(2023)Wu, Bansal, Zhang, Wu, Li, Zhu, Jiang, Zhang, Zhang,
  Liu, Awadallah, White, Burger, and Wang]{autogen}
Qingyun Wu, Gagan Bansal, Jieyu Zhang, Yiran Wu, Beibin Li, Erkang Zhu,
  Li~Jiang, Xiaoyun Zhang, Shaokun Zhang, Jiale Liu, Ahmed~Hassan Awadallah,
  Ryen~W White, Doug Burger, and Chi Wang.
\newblock Autogen: Enabling next-gen llm applications via multi-agent
  conversation, 2023.
\newblock URL \url{https://arxiv.org/abs/2308.08155}.

\bibitem[Zhu et~al.(2025)Zhu, Du, Hong, Yang, Guo, Wang, Wang, Qian, Tang, Ji,
  and You]{multiagentbench}
Kunlun Zhu, Hongyi Du, Zhaochen Hong, Xiaocheng Yang, Shuyi Guo, Zhe Wang,
  Zhenhailong Wang, Cheng Qian, Xiangru Tang, Heng Ji, and Jiaxuan You.
\newblock Multiagentbench: Evaluating the collaboration and competition of llm
  agents, 2025.
\newblock URL \url{https://arxiv.org/abs/2503.01935}.

\end{thebibliography}

\newpage
\appendix
\section{Formal Model Details}
\label{app:formal_details}

\subsection{Edit-Set Algebra and SEC}
\label{app:algebra}

The workspace state $\mathcal{W}_t$ maps each file to its contents at time $t$; each agent keeps a local replica $\mathcal{W}_t^{i}$ that our substrate keeps consistent with its peers.
An \emph{edit} inserts or deletes a payload string at a position in one file. Token-level edits, whose payload is one or more BPE/SentencePiece tokens, are the natural patch granularity an LLM emits. Because every token is a contiguous string of characters, token-level edits embed losslessly into the character-level Yrs alphabet~\citep{yjs,shapiro_crdt}, so the substrate accepts LLM-generated patches without re-encoding.

A standard LLM coding agent, working alone, produces a \emph{patch sequence}. It samples each patch $P_t$ from $\pi(\cdot \mid \mathcal{W}_{t-1})$, conditioned on the prior workspace, and applies it in strict causal order. A concurrent multi-agent system instead produces an \emph{edit set}. Each agent $i$ samples its patches $P_t^{i}$ in parallel from its own replica, and our substrate folds them into a single state $\mathcal{W}_T = \textsc{merge}(\{P_t^{i}\}_{i,t})$. Because \emph{merge} is commutative and associative under the op-based sequence CRDT with Lamport timestamps, the workspace satisfies strong eventual consistency: any two replicas that have observed the same set of operations hold the same state.

\subsection{Information Density}
\label{app:infodensity}

Treat each patch as a random variable over the rubric coverage $X \in \{0,1\}^{D}$, one bit per scoring dimension, so that $I(P; X)$ measures how much a single patch contributes to the rubric. A sequential agent's patches add up by the chain rule, and because a competent LLM avoids redundant edits, each further patch adds less once the easy rubric bits are covered. In the concurrent setting with coordination, $N$ agents sample at step $t$ from the same $\mathcal{W}_t$ but are \emph{pushed onto disjoint files} by \texttt{claim}, so their patches are close to independent given the workspace, and the per-step joint information across agents
\begin{equation}
I\!\big(\{P_t^{i}\}_{i=1}^{N};\,X \mid \mathcal{W}_t\big) \;=\; \sum_{i=1}^{N} H(P_t^{i} \mid \mathcal{W}_t) \;-\; H\!\big(\{P_t^{i}\}_{i=1}^{N} \mid \mathcal{W}_t,\,X\big)
\label{eq:info}
\end{equation}
scales near-linearly in $N$ as long as claims keep the joint entropy from collapsing. Without coordination the $N$ agents collide on the same files, their patches become redundant, and the novel rubric-relevant information per wall-clock step stays near the single-agent rate, so the room wastes concurrent compute on redundant edits. This is the information-theoretic reading of the six-condition ablation: AgentRoom delivers higher quality not because individual agents are smarter, but because the coordination layer keeps per-step novel information from collapsing under collision.

\textbf{What the probe realises.} The bundle component probe (\S\ref{sec:discussion}, Appendix~\ref{app:bundle_probe}) realises this substrate--coordination decomposition, and both of its steps carry wide intervals at these $n$. Eq.~\ref{eq:collision} explains the asymmetry, since SEC alone leaves $p_{\text{collision}}$ untouched, while Eq.~\ref{eq:info} predicts that the gain should track rubric dimensionality $D$ and per-agent novelty rate, consistent with the larger absolute lifts observed on T4, with its larger multi-domain rubric, than on T1 (Appendix~\ref{app:more_results}).

\section{Additional Results}
\label{app:more_results}

\subsection{12-Stratum CMH Forest}
\label{app:forest}

\begin{figure}[h]
\centering
\includegraphics[width=\linewidth]{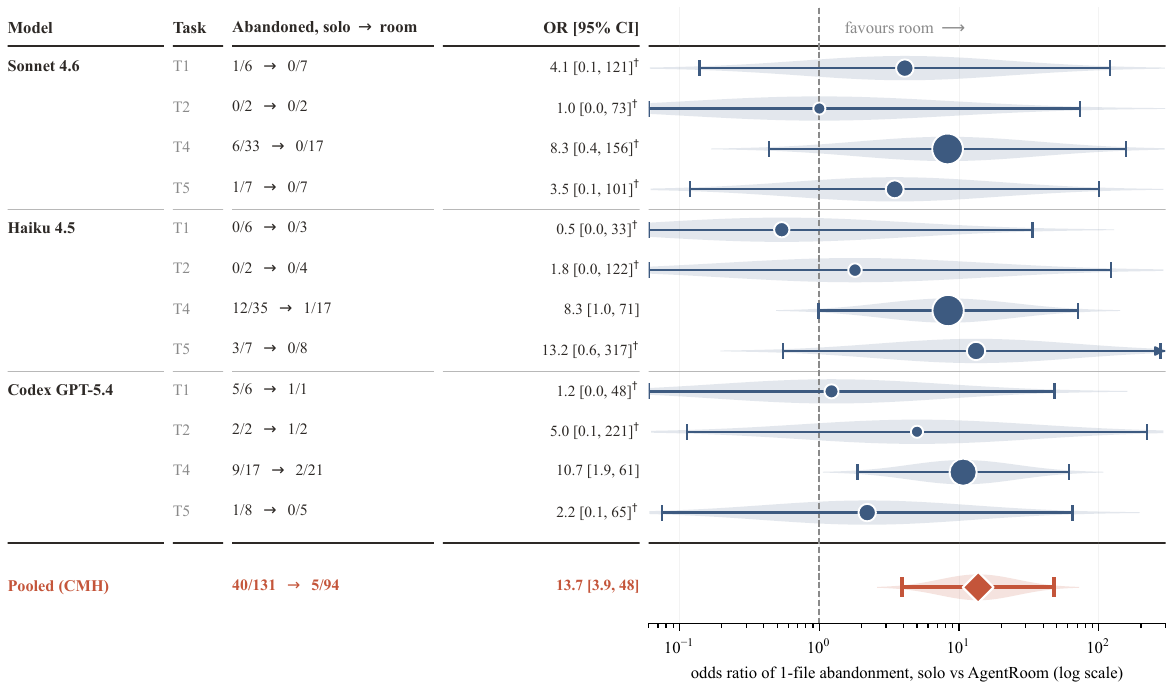}
\caption{Cochran-Mantel-Haenszel forest plot of Solo-vs-AgentRoom $1$-file abandonment across all $12$ model$\times$task strata (Sonnet/Haiku/Codex $\times$ T1/T2/T4/T5).}
\label{fig:forest}
\end{figure}

Figure~\ref{fig:forest} shows the abandonment-rate CMH analysis in \S\ref{sec:variance}---one row per stratum, marker area proportional to stratum size, whisker spanning the $95\%$ Wald interval on the log odds ratio. Each stratum with abandonment events points above OR$=1$; the pooled common odds ratio is $13.7$ ($95\%$ CI $[3.9,\,48]$, $\chi^2{=}22.8$, $p<10^{-5}$, Tarone homogeneity $p{=}0.92$), shown on the bottom row. Zero-event strata sit at OR$\approx 1$ and are marked $\dagger$, where we apply a Haldane $0.5$ correction so the interval is defined at all.

\subsection{Cross-Model ES (Tier III, Exploratory)}

\begin{figure}[h]
\centering
\includegraphics[width=0.85\linewidth]{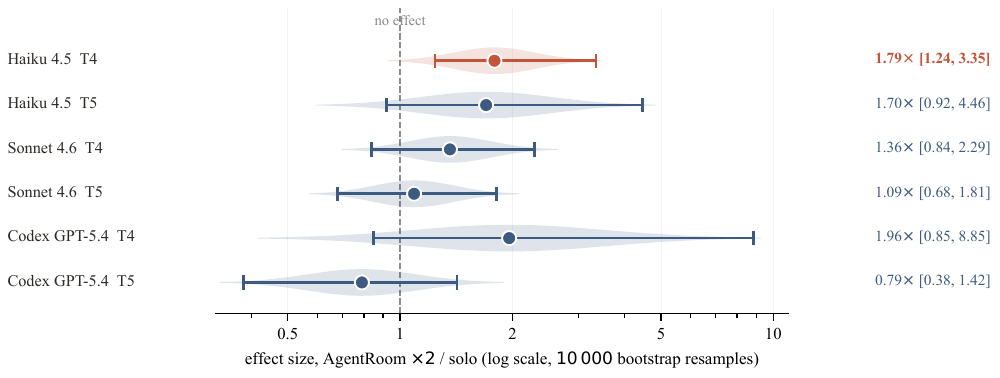}
\caption{Bootstrap effect-size distributions ($10\,000$ resamples) for the Tier III cross-model cells. Only Haiku T4 has its $95\%$ CI strictly above $1$; the other five cells span $0.79$--$1.96\times$ point-estimate with CIs that include $1$.}
\label{fig:bootstrap_es}
\end{figure}

We report the cross-model effect size of $\times 2$ over solo as exploratory only (Tier III; the variance and abandonment results carry the empirical claims). Table~\ref{tab:crossmodel} reports the numbers; Figure~\ref{fig:bootstrap_es} plots the same data. Only Haiku on T4 clears its 95\% bootstrap CI under the AST scorer (and its point estimate is positive under regex); we leave these results in the appendix to avoid seeming to lean on a result that the data do not support.

\begin{table}[h]
\centering
\caption{Cross-model T4 ES with bootstrap 95\% CI (10\,000 resamples), AST-scored subsample of the full pool ($\geq 30$\,s elapsed). $n$ differs from the LLM-scored pools because the AST scorer covers a subsample of the full pool. Headline scorer is AST (TypeScript compiler API); regex-scorer values after the slash for cross-check. $^\ast$ CI excludes 1. Two CLI-unstable models (Gemini, GPT-5.4-mini) excluded (see \S\ref{sec:cli_fragility}). Reported as exploratory conjecture: only Haiku T4 clears the CI under the AST scorer.}
\label{tab:crossmodel}
\small
\begin{tabular}{lccccc}
\toprule
Model & SWE-bench V. & Solo $(n)$ & $\times$2 $(n)$ & ES (AST / regex) & 95\% CI (AST) \\
\midrule
Haiku 4.5 & 73.3\% & 0.520 (13) & \textbf{0.929} (8) & \textbf{1.79$\times$} / 1.58$\times$ & [1.24, 3.35]$^\ast$ \\
Codex GPT-5.4 & $\sim$80\% & 0.248 (9) & 0.487 (13) & 1.96$\times$ / 1.07$\times$ & [0.85, 8.85] \\
Sonnet 4.6 & 79.6\% & 0.463 (11) & 0.630 (15) & 1.36$\times$ / 1.14$\times$ & [0.84, 2.29] \\
\bottomrule
\end{tabular}
\end{table}

\subsection{Scaling on T4 Sonnet}

Figure~\ref{fig:scaling} shows the agent-count sweep on T4 (Sonnet 4.6, budget-fair $30$--$700$\,s, LLM-judge composite), with every point labelled with its own $n$. The two scorers disagree on the optimum: quality as judged peaks at $\times 2$ while the mean number of passing tests peaks at $\times 3$ (\S\ref{sec:scaling}).

\subsection{Difficulty Gradient (T1/T2/T3/T4/T5, Sonnet 4.6)}

\begin{table}[!ht]
\centering
\caption{Sonnet 4.6 ES across the difficulty gradient (regex/AST scorer, outlier-filtered $\geq 0.3$).}
\label{tab:difficulty}
\small
\begin{tabular}{lccc}
\toprule
Task & Solo (n) & $\times 2$ (n) & ES \\
\midrule
T1 & 0.919 (4) & 0.910 (7) & 0.99$\times$ \\
T2 & 0.840 (3) & 0.904 (4) & 1.08$\times$ \\
T3 & 0.887 (3) & 0.943 (4) & 1.06$\times$ \\
T4 & 0.796 (10) & \textbf{0.958} (10) & \textbf{1.20}$\times$ \\
T5 & 0.697 (7) & 0.810 (7) & 1.16$\times$ \\
\bottomrule
\end{tabular}
\end{table}
Table~\ref{tab:difficulty} reports the ES of Sonnet 4.6 across all five tasks under the regex/AST scorer family (outlier-filtered, score $\geq 0.3$). We applied our LLM-judge rubric only to the T4 budget-fair pool, our headline cell. For consistency across tasks we retain the regex/AST scorer used the same way on all five tasks.
At low difficulty (T1, a single auth route), $\times 2$ saturates at the test-suite ceiling; medium-difficulty tasks (T2, T3) gain slightly. On the hard tasks, T4 and T5 gain most under the regex scorer (Table~\ref{tab:difficulty}), in the same direction as the LLM-judge T4 cell (Table~\ref{tab:ablation}).

\subsection{Heterogeneous Pairs (T4, Sonnet 4.6 partner)}

\begin{table}[!ht]
\centering
\caption{Heterogeneous vs.\ homogeneous $\times 2$ on T4 (LLM-judge, budget-fair). Tests is the mean over all budget-fair runs in the cell ($n{=}17$/$4$/$3$), which is not the LLM-scored $n$ shown in the Score column.}
\label{tab:heterogeneous}
\small
\begin{tabular}{lcc}
\toprule
Pair & Score (n) & Tests \\
\midrule
Sonnet $\times$2 (homo) & 0.669 (14) & \textbf{32.8} \\
Sonnet+Codex & \textbf{0.721} (3) & 16.0 \\
Sonnet+Gemini & 0.542 (3) & 22.0 \\
\bottomrule
\end{tabular}
\end{table}
Table~\ref{tab:heterogeneous} compares paired Sonnet with two heterogeneous pairings on T4. Sonnet+Codex scores the highest on the LLM-judge composite at $0.721$ ($n{=}3$), but is underpowered relative to the homogeneous Sonnet$\times 2$ baseline ($n{=}14$), and our two scorers disagree on it: the mean number of passing tests is lower for both heterogeneous pairings than for homogeneous Sonnet$\times 2$. We treat the heterogeneous pairings as exploratory.

\subsection{Cost-Equivalent Paired Model (Tier II.b, T4)}
\label{app:cost}

\begin{figure}[h]
\centering
\includegraphics[width=0.80\linewidth]{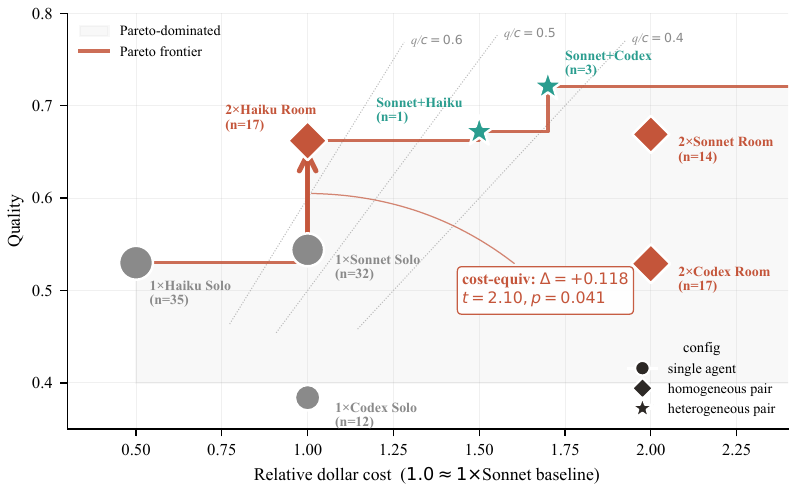}
\caption{Cost--quality Pareto on T4. Shaded region is Pareto-dominated; dotted lines are iso-efficiency (quality/cost) contours. $2{\times}$Haiku AgentRoom clears $1{\times}$Sonnet at matched plotted cost.}
\label{fig:cost_pareto}
\end{figure}

At no more than the dollar cost of one Sonnet agent on T4 (Figure~\ref{fig:cost_pareto}), under our LLM-judge rubric, $2{\times}$Haiku AgentRoom reaches mean $0.662$ ($n{=}17$, $\sigma{=}0.165$) versus $1{\times}$Sonnet $0.544$ ($n{=}32$, $\sigma{=}0.222$)---Welch $t{=}2.10$, two-sided $p{=}0.041$.
Haiku~4.5 is $3\times$ cheaper per token than Sonnet~4.6, so two Haiku agents cost about two-thirds as much as one Sonnet agent.
We see this as a Pareto-competitive point for tasks of T4's difficulty only.
The score gap is moderate, and on tasks where solo Haiku already collapses we do not expect it to hold: the Codex T5 ES regression (\S\ref{sec:results}) suggests that splitting the work reduces a pair's gain.

\subsection{Compute Cost (T4, Sonnet 4.6)}
\label{app:compute}

Table~\ref{tab:cost} separates wall-clock (elapsed) from cumulative compute (sum of per-agent CLI durations, a proxy for billable cost). AgentRoom runs share the wall-clock budget concurrently, so wall stays close to solo while compute scales with $N$.
The AgentRoom $\times 2$ cell sits at the top of the quality column at the same wall-clock budget as the solo and parallel-merge baselines; and the $\times 3$ cell pays an additional $\sim 500$\,s of compute for a small quality regression.

\begin{table}[h]
\centering
\caption{Wall vs compute on T4 Sonnet (LLM-judge composite, budget-fair pool).}
\label{tab:cost}
\small
\begin{tabular}{lcccc}
\toprule
Condition & $N$ & Wall (s) & Compute (s) & Quality \\
\midrule
Solo & 1 & 550 & 550 & 0.544 \\
Parallel merge & 2 & 558 & 1077 & 0.456 \\
Shared only & 2 & 601 & 1201 & 0.575 \\
AgentRoom $\times$2 & 2 & 561 & 1072 & \textbf{0.669} \\
AgentRoom $\times$3 & 3 & 566 & 1572 & 0.553 \\
\bottomrule
\end{tabular}
\end{table}

\subsection{T5 Cross-Model Matrix}
\label{app:t5_matrix}

T5, the largest task in the suite (algorithmic trading platform, 900\,s budget), exposes the per-model architectural-preference interaction discussed in \S\ref{sec:results}. Table~\ref{tab:crossmodel_t40} reports the per-model cells under the AST scorer: Haiku $\times 2$ has the only large point estimate ($1.70\times$), Sonnet $\times 2$ is flat, and Codex $\times 2$ regresses (consistent with the model's monolithic-architecture default resisting a split across two agents). We omit GPT-5.4-mini, which produced a single valid solo run ($0.077$) before its CLI crashed (\S\ref{sec:cli_fragility}).
The heterogeneous and mixed T5 cells are one to four runs each, so we list them here rather than tabulate them, for completeness only: Sonnet+GPT-5.4 $0.857$ ($n{=}2$), Sonnet+Haiku $0.900$ ($n{=}1$), Sonnet+Gemini $0.626$ ($n{=}3$), Mixed S+C+G $\times 3$ $0.673$ ($n{=}4$), and Mixed S+H+C $\times 3$ $0.791$ ($n{=}2$).

\begin{table}[h]
\centering
\caption{T5 (trading platform, 900\,s budget), AST scorer, per-model cells. All CLI-stable cells are point-estimate-positive on $\times 2$ except Codex; CIs do not exclude 1 at this $n$. The Gemini row is CLI-unstable and informational only (\S\ref{sec:cli_fragility}); sparse heterogeneous cells are in the text above.}
\label{tab:crossmodel_t40}
\begin{tabular}{lccccc}
\toprule
Model & Solo $(n)$ & $\times$2 $(n)$ & $\times$3 $(n)$ & ES($\times$2) & Tests ($\times$2) \\
\midrule
Haiku 4.5 & 0.416 (8) & \textbf{0.707} (7) & 0.822 (1) & \textbf{1.70$\times$} & 38 \\
Sonnet 4.6 & 0.618 (7) & 0.674 (7) & 0.730 (2) & 1.09$\times$ & 57 \\
GPT-5.4 & 0.667 (9) & 0.531 (7) & 0.761 (1) & 0.79$\times$ & 8 \\
Gemini Flash \emph{(unstable)} & 0.851 (3) & 0.506 (9) & 0.246 (3) & 0.59$\times$ & 12 \\
\bottomrule
\end{tabular}
\end{table}

\subsection{Behavior Frequency (T4 $\times$2, Sonnet 4.6, $n{=}6$)}

We coded each of six valid $\times 2$ AgentRoom runs on T4 Sonnet for six categories of collaborative behavior (Table~\ref{tab:behavior}). 

Almost every run claims a module and adjusts the plan---cross-agent bug fix and apology appear in only a third of runs and are not the typical case. 

The protocol asks only that agents broadcast claims and respect the claims of others; the apologies, gap-filling, and explicit ``do not'' warnings visible in the transcripts of Appendix~\ref{app:room-logs} are not in the prompt.

\begin{table}[h]
\centering
\caption{Frequency of collaborative behaviors across $6$ valid $\times 2$ AgentRoom runs on T4 Sonnet.}
\label{tab:behavior}
\begin{tabular}{lc}
\toprule
Behavior & Runs exhibiting (of 6) \\
\midrule
Module claiming via broadcast & 6/6 \\
Plan adjustment after reading teammate & 5/6 \\
Export list shared for teammate & 4/6 \\
Proactive gap filling & 3/6 \\
Cross-agent bug fix & 2/6 \\
Apology for file boundary violation & 2/6 \\
\bottomrule
\end{tabular}
\end{table}

\subsection{Pool-Wide Claim Coverage (T4, all models)}
\label{app:claim_coverage}

The behavioral coding above is a hand annotation over six runs. To count protocol adherence pool-wide, we replayed the archived room logs of every scored budget-fair T4 AgentRoom run whose room log contains agent-authored coordination text ($n{=}67$, including the $\times 3$/$\times 4$ scaling runs). The remaining scored runs coordinated through MCP tool state the archiver did not capture, so their claims are unobservable post hoc. An LLM extractor---the same headless mechanism as the judge of \S\ref{sec:scoring}---parses each room's free-text claims into per-agent file, directory, and blanket claims, cached per run for audit. Each changed source file in the run's archived file-change record then matches deterministically against the union of all claims. A changed file that no agent ever claimed is an \emph{unclaimed write}---the automatable half of the protocol-violation count. The archived runs carry no per-agent authorship (agents share one workspace), so the finer rate of edits to a teammate's claimed file is not recoverable from the archive.

Across the $65$ runs with at least one changed source file, the mean unclaimed-write rate is $0.11$, and $39/65$ runs have every changed file covered by a claim. The rate is uncorrelated with the number of changed files ($r{=}0.02$), so it is not a repository-size artifact. Against the LLM-judge composite the unclaimed-write rate correlates positively---Pearson $r{=}0.28$ (bootstrap 95\% CI $[0.06, 0.46]$), Spearman $\rho{=}0.30$ $[0.06, 0.52]$; restricted to $\times 2$ runs, $r{=}0.24$ ($n{=}54$). Runs with zero unclaimed writes average $0.58$ composite against $0.67$ for runs with at least one.

\textbf{How to read it.} Deviating from claim discipline therefore does not predict lower quality in this pool; if anything, runs containing unclaimed writes score modestly higher. This is the quantitative counterpart of the advisory-lock design (\S\ref{sec:protocol}): the room surfaces a violation rather than blocking it, and the transcripts behind Table~\ref{tab:behavior} show that the unclaimed write is often the unblocking edit. The estimate is correlational, zero-inflated, and conditioned on rooms whose coordination went through the observable channel.

\subsection{Cross-Model Ablation (T4, all CLI-stable models)}

To test whether the substrate-vs-AgentRoom separation (Table~\ref{tab:ablation}) holds beyond our single (Sonnet, T4) cell, we also ran the parallel-merge, shared-only, and noMCP conditions on Haiku 4.5 and on Codex GPT-5.4, as Table~\ref{tab:ablation_xmodel} reports under the LLM-judge rubric (on the budget-fair pool, $30$--$700$\,s).
For Sonnet and Haiku, AgentRoom sits at the top of the model's column; for both, the parallel-merge cell falls below the solo baseline, replicating the ``naive concurrent ensembling underperforms a single agent'' finding from Sonnet.
The intermediate cells are noisier on Haiku: its noMCP cell ($0.579$, $n{=}6$) is just below shared-only ($0.657$, $n{=}7$), not above it as on Sonnet, so for Haiku we read only the AgentRoom-vs-baseline endpoints, not the fine-grained substrate-then-prompt step ordering, which is identified only on the larger Sonnet cells. Codex differs at the AgentRoom cell. Pairs of Codex agents do worse than the other AgentRoom pairs, though better than one Codex agent, which fits the Codex T5 regression and the model's strong default toward monolithic single-file architecture (Appendix~\ref{app:t5_matrix}).

\begin{table}[h]
\centering
\caption{Cross-model T4 ablation (LLM-judge composite, budget-fair pool). For Sonnet and Haiku, AgentRoom is the top condition and parallel-merge falls below solo. Codex non-Solo/AgentRoom cells were not collected (vendor-CLI hang).}
\label{tab:ablation_xmodel}
\small
\begin{tabular}{lcccc!{\vrule width 0.8pt}c}
\toprule
& \multicolumn{4}{c!{\vrule width 0.8pt}}{\textit{Baselines and partial bundles}} & \multicolumn{1}{c}{\textit{Ours}} \\
\cmidrule(lr){2-5} \cmidrule(lr){6-6}
Model & Solo & Parallel merge & Shared only & noMCP & AgentRoom $\times 2$ \\
\midrule
Sonnet 4.6 & 0.544 ($n{=}32$) & 0.456 ($n{=}12$) & 0.575 ($n{=}11$) & 0.588 ($n{=}7$) & \textbf{0.669} ($n{=}14$) \\
Haiku 4.5  & 0.530 ($n{=}35$) & 0.383 ($n{=}12$) & 0.657 ($n{=}7$)  & 0.579 ($n{=}6$) & \textbf{0.662} ($n{=}17$) \\
Codex GPT-5.4 & 0.384 ($n{=}12$) & -- & -- & -- & 0.529 ($n{=}17$) \\
\bottomrule
\end{tabular}
\end{table}

\subsection{Codebook Threshold Sensitivity (Tier I CMH)}
\label{app:codebook_sensitivity}

\paragraph{Implementation.} We implement the codebook as a deterministic Python script that reads each run's recorded duration, the regex-scorer composite of its snapshot, and the snapshot's source-file count, and emits a label without human judgment. We committed the script to the paper repository before the CMH analyses. Because classification is deterministic, blinded inter-rater agreement is trivially $\kappa{=}1$; residual classification risk lies in the threshold choice, which we sweep below.

\textbf{The pre-stated codebook.} The pre-stated codebook for failure-mode classification (\S\ref{sec:variance}) labels a $<0.3$ run as \emph{1-file abandonment} if duration $<200$\,s OR source files $\leq 2$, and as \emph{infrastructure failure} otherwise.
Table~\ref{tab:codebook_sweep} sweeps both thresholds and reports the resulting CMH common odds ratio for the T4-only analysis (3 model strata) under each threshold combination. Across all 12 threshold combinations the T4-only CMH OR is at least $6.3$ and CMH $p<0.001$, so the Tier~I finding does not depend on a sharp threshold choice.
Moving the duration cutoff within $150$--$300$\,s does not shift any cell counts (every $<0.3$ abandonment in our pool either exited within $150$\,s or ran the full budget); the source-file cutoff matters more (more permissive thresholds reclassify Codex AgentRoom runs from infrastructure to abandonment, lowering OR but never below $6.3$).
The 12-stratum CMH (3 models $\times$ 4 tasks, OR$=13.7$) reported in \S\ref{sec:variance} is correspondingly stable under the threshold sweep.

\begin{table}[h]
\centering
\caption{Threshold sensitivity for the Tier I CMH. Each row is a threshold combination; cells report Solo/AgentRoom abandonment counts per model and the resulting CMH common OR with $95\%$ CI and $p$-value.}
\label{tab:codebook_sweep}
\small
\resizebox{\linewidth}{!}{%
\begin{tabular}{cc!{\vrule width 0.8pt}cccccc!{\vrule width 0.8pt}ccc}
\toprule
\multicolumn{2}{c!{\vrule width 0.8pt}}{\textit{Thresholds}} &
\multicolumn{2}{c}{Sonnet 4.6} & \multicolumn{2}{c}{Haiku 4.5} &
\multicolumn{2}{c!{\vrule width 0.8pt}}{Codex GPT-5.4} &
\multicolumn{3}{c}{\textit{CMH pooled}} \\
\cmidrule(lr){1-2} \cmidrule(lr){3-4} \cmidrule(lr){5-6} \cmidrule(lr){7-8} \cmidrule(lr){9-11}
$d$\,thr & $f$\,thr & Solo & Room & Solo & Room & Solo & Room & OR & 95\% CI & $p$ \\
\midrule
$<150$ & $\leq 1$ & 6/33 & 0/17 & 12/35 & 1/17 & 9/17 & 0/21 & 24.2 & [3.5, 169] & 0.00001 \\
$<150$ & $\leq 2$ & 6/33 & 0/17 & 12/35 & 1/17 & 9/17 & 2/21 & 11.9 & [3.0, 47] & 0.00005 \\
$<150$ & $\leq 3$ & 6/33 & 0/17 & 12/35 & 1/17 & 10/17 & 6/21 & 6.3 & [2.1, 19] & 0.00069 \\
$<200$ & $\leq 1$ & 6/33 & 0/17 & 12/35 & 1/17 & 9/17 & 0/21 & 24.2 & [3.5, 169] & 0.00001 \\
$<200$ & $\leq 2$ \emph{(paper)} & \textbf{6/33} & \textbf{0/17} & \textbf{12/35} & \textbf{1/17} & \textbf{9/17} & \textbf{2/21} & \textbf{11.9} & \textbf{[3.0, 47]} & \textbf{0.00005} \\
$<200$ & $\leq 3$ & 6/33 & 0/17 & 12/35 & 1/17 & 10/17 & 6/21 & 6.3 & [2.1, 19] & 0.00069 \\
$<250$ & $\leq 1$ & 6/33 & 0/17 & 12/35 & 1/17 & 9/17 & 0/21 & 24.2 & [3.5, 169] & 0.00001 \\
$<250$ & $\leq 2$ & 6/33 & 0/17 & 12/35 & 1/17 & 9/17 & 2/21 & 11.9 & [3.0, 47] & 0.00005 \\
$<250$ & $\leq 3$ & 6/33 & 0/17 & 12/35 & 1/17 & 10/17 & 6/21 & 6.3 & [2.1, 19] & 0.00069 \\
$<300$ & $\leq 1$ & 6/33 & 0/17 & 12/35 & 1/17 & 9/17 & 0/21 & 24.2 & [3.5, 169] & 0.00001 \\
$<300$ & $\leq 2$ & 6/33 & 0/17 & 12/35 & 1/17 & 9/17 & 2/21 & 11.9 & [3.0, 47] & 0.00005 \\
$<300$ & $\leq 3$ & 6/33 & 0/17 & 12/35 & 1/17 & 10/17 & 6/21 & 6.3 & [2.1, 19] & 0.00069 \\
\bottomrule
\end{tabular}%
}
\end{table}

\subsection{Bundle Component Probe (CRDT + prompt, no MCP tools)}
\label{app:bundle_probe}

\textbf{The probe condition.} A natural question is whether the AgentRoom bundle's components (CRDT layer + MCP coordination tools + collaboration prompt) can be partially separated.
To find out whether they can, we ran an additional condition that keeps the CRDT-merged shared workspace and the collaboration prompt but removes the MCP tool surface (no \texttt{room\_claim}/\texttt{room\_broadcast}, no MCP server).
Sonnet 4.6, T4, $1200$\,s budget, $30$--$700$\,s budget-fair envelope, $n{=}7$; LLM-judge mean $0.588$ ($\sigma{=}0.188$).
\textbf{Where the cell sits.} The cell sits between shared-only ($0.575$, no prompt, no tools) and AgentRoom ($0.669$, with both); the collaboration prompt above the shared CRDT substrate adds only $+0.013$ over shared-only on its own, while the MCP tool surface adds a further $+0.081$. At these cell sizes the interval on the MCP-tool step spans zero (see the ordering claim in \S\ref{sec:discussion}). The split is also sensitive to the pool boundary; widening the elapsed-time window moves it toward the substrate, which is the second reason we report the ordering rather than a percentage.
The component contributions are not strictly additive across conditions: the prompt-only Sonnet control without teammates falls below bare solo (Appendix~\ref{app:prompt_control}), so the prompt's contribution is sign-dependent on the substrate.

\subsection{Solo+Collab-Prompt Control (T4, Sonnet and Haiku)}
\label{app:prompt_control}

To isolate prompt-vs-channel effects we ran solo with a structurally similar collaboration prompt (same tool list, teammate references stripped).
The dependence on the model (Table~\ref{tab:prompt_control}) is informative.

\begin{table}[h]
\centering
\caption{Solo + collaboration-style prompt without teammates, T4, $n{=}3$ each (LLM-judge composite). The structured prompt alone is model-dependent: it hurts Sonnet (channel is the load-bearing variable) and helps Haiku (the structured prompt also acts as workflow scaffolding); the channel and prompt-as-workflow are not separately identified for Haiku at this $n$.}
\label{tab:prompt_control}
\small
\begin{tabular}{lccc}
\toprule
Model & Solo (bare, budget-fair) & Solo + collab prompt ($n{=}3$) & AgentRoom $\times 2$ (budget-fair) \\
\midrule
Sonnet 4.6 & 0.544 ($n{=}32$) & 0.448 (range 0.20--0.62) & \textbf{0.669} ($n{=}14$) \\
Haiku 4.5  & 0.530 ($n{=}35$) & \textbf{0.604} (range 0.55--0.66) & 0.662 ($n{=}17$) \\
\bottomrule
\end{tabular}
\end{table}

For Sonnet, the structured prompt alone scores below bare solo and far below AgentRoom $\times 2$, so the AgentRoom gain is attributable to the channel rather than to prompt complexity.
One Sonnet control run scored at the bottom of the range because the agent burned the wall-clock budget calling room tools that returned no useful state; without a multi-agent context to refer to, the prompt becomes counterproductive.
For Haiku, the same prompt scores above bare solo and is statistically indistinguishable from AgentRoom $\times 2$ at this $n$.
We attribute this to Haiku's stronger benefit from workflow scaffolding (consistent with the lone-agent abandonment failure mode that the explicit tool list defuses); for Haiku, this control does not separate the channel from the prompt-as-workflow.

\subsection{Cross-Domain Validation: DevBench (multi-file Python from PRD)}
\label{app:devbench}

We ran 10 DevBench~\citep{devbench} Python projects with Haiku 4.5 in solo and $\times$2 AgentRoom conditions at a 600\,s wall-clock budget.
Each task comes with a PRD, an architecture sketch and a hidden \texttt{pytest} suite, and requires implementing the multi-file project from scratch (we give the agents no scaffolded source code).

\textbf{Pass rates.} The pass rate (Table~\ref{tab:devbench}): $\times$2 AgentRoom passes 8 of 10 tasks, and solo passes 7 of 10.
Two tasks (\texttt{stocktrends}, \texttt{textcnn}) fail under both conditions, indicating task difficulty above what 600\,s of Haiku affords regardless of agent count.
The 10 concurrent runs had zero CRDT conflicts except a single benign 1-merge event on \texttt{lice} that did not affect correctness. At $n{=}10$ a one-task pass differential carries no inferential weight, and we report this as a descriptive cross-domain check rather than a quantitative claim.
The cross-domain evidence is consistent with our TypeScript results: $\times$2 does not break correctness on multi-file Python projects, and on one task, \texttt{geotext}, the $\times$2 condition finished in 390\,s while solo timed out at 603\,s, which accounts for the +1 difference.

\begin{table}[h]
\centering
\caption{DevBench cross-domain validation (Haiku 4.5, 600\,s budget). \emph{Pass} = hidden pytest suite passes; durations in seconds. \texttt{geotext}: $\times 2$ completes within budget while solo times out; \texttt{stocktrends, textcnn}: both fail (task too hard for budget).}
\label{tab:devbench}
\begin{tabular}{lcc}
\toprule
Task & Solo (PASS, dur) & $\times$2 AgentRoom (PASS, dur) \\
\midrule
\texttt{arxiv-digest} & PASS, 70\,s & PASS, 120\,s \\
\texttt{readtime} & PASS, 110\,s & PASS, 100\,s \\
\texttt{chakin} & PASS, 130\,s & PASS, 130\,s \\
\texttt{geotext} & \emph{FAIL, 603\,s timeout} & \textbf{PASS, 390\,s} \\
\texttt{hone} & PASS, 90\,s & PASS, 160\,s \\
\texttt{hybrid-images} & PASS, 100\,s & PASS, 170\,s \\
\texttt{lice} & PASS, 330\,s & PASS, 270\,s \\
\texttt{particle-swarm-opt} & PASS, 100\,s & PASS, 120\,s \\
\texttt{stocktrends} & FAIL, 600\,s & FAIL, 480\,s \\
\texttt{textcnn} & FAIL, 601\,s & FAIL, 856\,s \\
\midrule
Pass rate & 7/10 & \textbf{8/10} \\
\bottomrule
\end{tabular}
\end{table}

\subsection{Cross-Language Validation: Rust + axum (T4 Transposed)}
\label{app:rust}

\spaceskip=\fontdimen2\font plus \fontdimen3\font minus 0.05\fontdimen4
\textbf{Setup.} To address the same-runtime concern (\S\ref{sec:limitations}), we ran the T4 fintech-ledger task transposed to Rust + axum~0.7 with Codex GPT-5.4 (solo and AgentRoom~$\times 2$). Same hidden test surface (\texttt{cargo test} on a vendor-pinned tokio/axum/serde stack), a $700$\,s wall-clock budget (the T4 budget-fair envelope top), the same LLM-judge composite. All $4$ AgentRoom~$\times 2$ runs completed cleanly under concurrent MCP execution with full GPT-5.4 (\S\ref{sec:cli_fragility} for the Codex-mini deployment caveat), and no \texttt{room\_claim} crashes.
\textbf{Cell means.} Cell means at the same LLM-judge composite as the TypeScript cells: Solo mean $0.714$ ($n{=}5$, $\sigma{=}0.104$)---AgentRoom~$\times 2$ mean $0.740$ ($n{=}4$, $\sigma{=}0.070$); $\Delta{=}{+}0.026$, in the same direction as the TypeScript headline (Welch underpowered at this $n$, $t{=}0.44$, $p{=}0.68$; we report descriptively). Code volume per AgentRoom~$\times 2$ run: $2008$--$2452$ Rust lines across $3$--$6$ \texttt{.rs} files. The replication establishes that the LLM-judge composite extends to Rust without breaking and that Codex MCP-concurrent execution is stable on full GPT-5.4.
One earlier AgentRoom~$\times 2$ run (score $0.580$) is excluded from this cell for a coordination artifact---the same four enums duplicated across two files with slightly different shapes; it is the only run in the Rust pool with this pattern, and it remains in the run archive.

\spaceskip=0pt\relax

\subsection{Execution-Oracle Replay (T4 Capability Probe)}
\label{app:oracle}

The scorers of \S\ref{sec:scoring} judge code but never execute it. As a check on the reported ordering, we replayed every archived T4 budget-fair snapshot against a task-blind capability probe fixed before the replay. The sandbox template plus the snapshot's \texttt{src/} overlay boots under \texttt{tsx}, and we probe ten binary capabilities against the live server.
We find that the oracle agrees with the judge on the endpoints of the ordering.
AgentRoom ranks first, and concurrency without coordination is at the floor: every parallel-merge snapshot boots but passes no endpoint capability, consistent with the silent-overwrite failure mode of \S\ref{sec:ablation}. The two scorers disagree on adjacent pairs in between: shared-only and noMCP sit within about $0.2$ capabilities of each other but in the opposite order to the judge's, so the probe corroborates only the endpoints of the \S\ref{sec:discussion} ordering; and ChatDev-style scores slightly above parallel-merge under the oracle, while the judge orders them the other way.
Mean capabilities passed, of ten (boot, accounts endpoint, account creation, account listing, balanced-transaction posting, unbalanced-transaction rejection, balance query, trial balance, audit trail, and reconciliation endpoint): AgentRoom~$\times 2$ $5.06$ ($n{=}17$), shared-only $4.91$ ($n{=}11$), shared+collab noMCP $4.70$ ($n{=}10$), Solo $3.67$ ($n{=}33$), ChatDev-style $1.46$ ($n{=}13$), parallel-merge $1.00$ ($n{=}15$).
Our probe pool covers all budget-fair runs, including runs without an LLM score, so $n$ differs slightly from Table~\ref{tab:ablation}.

\section{Full AgentRoom Logs}
\label{app:room-logs}

We drew the two transcripts below from each run's archived room log (\texttt{room\_log.json}), condensed for layout; wording, casing, field labels such as \textbf{CLAIMING}, self-chosen speaker names, and punctuation inside messages, including dashes, are the agents' own. Timestamps, speaker headers, and list markup are typeset from the log structure, with speaker colors following Figure~\ref{fig:dialogue}, which pairs a single representative run's chat broadcasts with the per-file ownership timeline.

\begin{figure}[h]
\centering
\includegraphics[width=\linewidth]{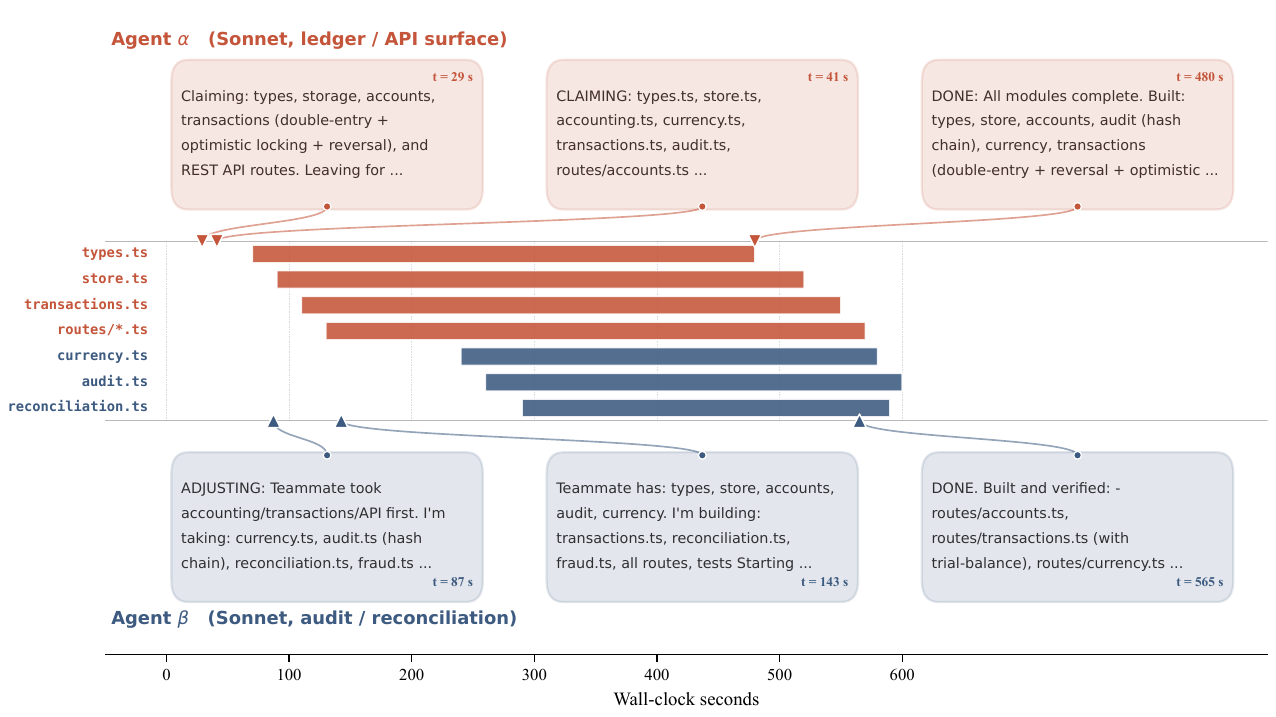}
\caption{Single representative T4 Sonnet AgentRoom~$\times 2$ run. Top swim-lane: agent $\alpha$ broadcasts (timestamps in-card); middle: per-file ownership Gantt (rust = $\alpha$, slate = $\beta$); bottom: agent $\beta$. Bezier S-curve leaders link each chat broadcast to its time on the Gantt. The whole protocol (claim, conflict-detect, hand-off, completion broadcast) emerges from the advisory six-step prompt (\S\ref{sec:protocol}) without any orchestrator.}
\label{fig:dialogue}
\end{figure}

\subsection{T4 $\times$3 Sonnet 4.6 : Complete AgentRoom Transcript}

\begin{roomlog}{Room broadcast log, T4 $\times$3 Sonnet 4.6, from \texttt{messages.jsonl}}
\small
\logmsg{cRust}{Agent-A}{47}
\textbf{CLAIMING:} \texttt{src/types.ts}, \texttt{src/store.ts}, \texttt{src/accounting.ts}, \texttt{src/currency.ts}, \texttt{tests/accounting.test.ts}\\
\textbf{STATUS:} working

\logmsg{cSlate}{Agent-B}{75}
\textbf{CLAIMING:} \texttt{src/transactions.ts}, \texttt{src/audit.ts}, \texttt{tests/transactions.test.ts}, \texttt{tests/audit.test.ts}\\
\textbf{STATUS:} in\_progress — waiting on types.ts from Dev A, will build concurrently

\logmsg{cRust}{Agent-A}{248}
\textbf{DONE:} \texttt{src/types.ts}, \texttt{src/store.ts}, \texttt{src/accounting.ts}, \texttt{src/currency.ts}, \texttt{tests/accounting.test.ts}\\
All 52 tests passing. Key exports for teammates:
\begin{itemize}[nosep,leftmargin=1.4em]
\item \texttt{src/types.ts}: Account, Transaction, AuditEntry, etc.
\item \texttt{src/store.ts}: singleton store, optimistic locking
\item \texttt{src/accounting.ts}: createAccount, postTransaction, reverseTransaction, generateTrialBalance
\item \texttt{src/currency.ts}: seedExchangeRates, convert, getRateAt
\end{itemize}
Developer C: \texttt{src/server.ts} needs seedExchangeRates()

\logmsg{cSlate}{Agent-B}{372}
\textbf{STATUS:} DONE — all 70 tests passing (35 new from B)\\
\textbf{NOTES:}
\begin{itemize}[nosep,leftmargin=1.4em]
\item \texttt{src/audit.ts}: AuditLog class (SHA-256 hash chain)
\item \texttt{src/transactions.ts}: TransactionStore (optimistic locking)
\end{itemize}
Dev C: import from \texttt{src/accounting.ts} for routes

\logmsg{cSlate}{Agent-B}{392}
Developer B now claiming: \texttt{src/reconciliation.ts}, \texttt{src/fraud.ts}, \texttt{src/routes.ts}, \texttt{src/server.ts} update

\logmsg{cSage}{Agent-C}{577}
\textbf{ACTIONS:}
\begin{itemize}[nosep,leftmargin=1.4em]
\item \textbf{BUG FIX} in \texttt{src/routes.ts}: \texttt{/accounts/path/:path(*)} invalid in Express 5. Fix: change to \texttt{/accounts/path/:path} (Dev B: I touched your file for a critical 1-line fix to unblock all tests — sorry!)
\item \textbf{CREATING:} \texttt{tests/api.test.ts} — comprehensive API integration tests covering all 8 spec requirements
\end{itemize}

\medskip\noindent\textbf{FINAL RESULT:} 98/100 tests passing, 10 source files
\end{roomlog}

\subsection{T4 $\times$4 Sonnet 4.6 : Conflict Resolution}

\begin{roomlog}{Room broadcast log, T4 $\times$4 Sonnet 4.6, from \texttt{messages.jsonl}}
\small
\logmsg{cRust}{DEV-A}{30}
\textbf{CLAIMING:} \texttt{src/types.ts}, \texttt{src/store.ts}, \texttt{src/accounts.ts}, \texttt{src/ledger.ts}, \texttt{src/transactions.ts}

\logmsg{cOchre}{Developer-3}{32}
\textbf{CLAIMING:} \texttt{src/types.ts}, \texttt{src/currency.ts}, \texttt{src/transactions.ts}\\
Module: Core types, multi-currency, transaction processing

\logmsg{cRust}{DEV-A}{45}
\textbf{CONFLICT RESOLUTION:} Dropping types.ts, transactions.ts\\
\textbf{REVISED CLAIM:} \texttt{src/store.ts}, \texttt{src/accounts.ts}, \texttt{src/ledger.ts}

\logmsg{cOchre}{Developer-3}{48}
\textbf{CONFLICT DETECTED} with DEV-A on types.ts, transactions.ts\\
\textbf{REVISED:} \texttt{src/currency.ts}, \texttt{src/audit.ts}, \texttt{src/fraud.ts}

\logmsg{cSlate}{DEV-B}{50}
\textbf{CLAIMING:} \texttt{src/audit.ts}, \texttt{src/fraud.ts}, \texttt{src/reconciliation.ts}, \texttt{src/routes.ts}\\
\textbf{NOTE:} DEV-A + Developer-3 conflict on types.ts + transactions.ts — please coordinate!

\logmsg{cOchre}{Developer-3}{60}
Re-claiming dropped files. \textbf{FINAL CLAIMS:} \texttt{src/types.ts}, \texttt{src/currency.ts}, \texttt{src/transactions.ts}, \texttt{src/audit.ts}, \texttt{src/fraud.ts}\\
DEV-A keeps: store, accounts, ledger

\logmsg{cSage}{DEV-C}{120}
\textbf{FINAL CLAIM} (4th developer, no conflicts): \texttt{src/reconciliation.ts}, \texttt{src/server.ts}, \texttt{tests/integration.test.ts}

\logmsg{cSage}{DEV-C}{180}
\textbf{REVISED:} DEV-B has reconciliation. Taking: \texttt{src/errors.ts}, \texttt{src/validators.ts}, \texttt{src/server.ts}

\logmsg{cSage}{DEV-C}{350}
\textbf{GAP FILL:} Developer-3 has not created transactions.ts yet. To avoid blocking, DEV-C is writing these uncreated files.

\logmsg{cRust}{DEV-A}{500}
\textbf{DONE:} All 66 tests passing. Modules ready for integration.

\logmsg{cSlate}{DEV-B}{520}
\textbf{DONE:} audit.ts, fraud.ts, reconciliation.ts\\
\textbf{ALERT} to DEV-C: DO NOT create audit.ts, fraud.ts, reconciliation.ts, routes.ts — already created by DEV-B! Reading DEV-A types.ts + errors.ts now before writing routes.ts.

\medskip\noindent\textbf{FINAL RESULT:} 65/66 tests passing, 13 source files, 1 CRDT conflict resolved, 14 room messages
\end{roomlog}

\section{Task Specifications}
\label{app:tasks}

All tasks start from the same Express.js~5 sandbox (Node 22, vitest 2, TypeScript 5.6, \texttt{type: module}) with a minimal \texttt{package.json}, \texttt{tsconfig.json}, and a stub \texttt{src/server.ts}.
\S\ref{sec:filter} lists the exact pinned versions.
Agents are told the test command, the expected file structure (where prescribed), and the requirements; they are not shown reference solutions.

\subsection{T1: JWT Authentication System}

Build a complete authentication system from scratch: \texttt{src/models/user.ts} (User interface with \texttt{id}, \texttt{email}, \texttt{passwordHash}, \texttt{createdAt}; a UserStore class with in-memory Map storage and \texttt{create}, \texttt{findByEmail}, \texttt{findById} methods; SHA-256 password hashing via \texttt{crypto.createHash}); \texttt{src/auth/jwt.ts} (\texttt{generateToken}, \texttt{verifyToken}, 1-hour expiry, base64 encoding, no external JWT library); \texttt{src/auth/middleware.ts} (Express middleware that extracts a Bearer token, verifies it, attaches the user to \texttt{req}, returns 401 on missing or invalid tokens); \texttt{src/routes/auth.ts} (\texttt{POST /auth/register}, \texttt{POST /auth/login}, \texttt{GET /auth/me}); \texttt{src/server.ts} (mount auth routes); and \texttt{tests/auth.test.ts} (seven test cases covering registration, duplicate detection, login success and failure, unauthenticated access, authenticated access, and expired-token handling).
TypeScript strict mode; ESM with \texttt{.js} import extensions; 300-second wall-clock budget.

\subsection{T2: Marketplace API}

Build a marketplace API with user accounts (register, login, profile update), product listings (title, description, price, category, stock count, seller user id), an order system (cart, checkout, status: \texttt{pending}, \texttt{confirmed}, \texttt{shipped}, \texttt{delivered}), a review system tied to verified orders, and product search with filters (title keyword, category, price range, minimum rating).
Endpoints must return 400, 401, 404, 409 as appropriate.
TypeScript strict mode; Express; in-memory storage; ESM with \texttt{.js} extensions.
No prescribed file structure: agents must decide how to organize modules, which makes this task sensitive to coordination in the multi-agent setting. 300-second wall-clock budget.

\subsection{T3: Collaborative E-commerce}

Build an e-commerce backend API with interconnected modules: User (register, login, profiles), Product (CRUD, search, stock management), Order (cart, checkout, stock validation, status tracking), and Review (purchase-verified ratings, average per product).
Full validation, error handling, and integration tests.
This task is the same shape as T2 but with tighter cross-module invariants (a review requires a delivered order; checkout decrements stock atomically), which stresses inter-agent agreement on shared schemas.

\subsection{T4: Fintech Ledger}

Build a financial ledger system with: double-entry accounting (every transaction creates debit and credit entries that must balance); multi-currency support (USD, EUR, GBP, JPY with historical exchange rates, amounts stored as integer cents); atomic transaction processing with optimistic locking and rollback; an immutable audit trail with SHA-256 hash chain for tamper detection; a reconciliation engine (match internal ledger against external statements); fraud detection rules (threshold amounts, rapid-fire transactions, new counterparties); a comprehensive REST API; and tests covering all invariants. 600-second wall-clock budget.

\section{Quality Scorer Details}
\label{app:scoring}
\label{app:scorer}

\paragraph{LLM-judge (primary scorer for T4 budget-fair).}
All continuous-score statistics in the main text on the T4 budget-fair pool are produced by an LLM-judge: a single Sonnet~4.6 invocation per run with a fixed rubric.
The judge receives the task spec, the test command, the test pass/fail status, and the post-run repository snapshot (truncated to 4\,KB per file and 30\,KB total) and returns four 0--100 scores assessing (a)~\emph{spec coverage} (how many spec requirements have visible implementation), (b)~\emph{correctness signals} (proper types, error handling, no obvious bugs), (c)~\emph{code quality} (separation of concerns, naming, idiomatic patterns), and (d)~\emph{test rigor} (edge-case coverage, meaningful assertions).
The composite is a fixed-weight average ($0.35$ spec coverage, $0.30$ correctness, $0.20$ code quality, $0.15$ test rigor); weights were fixed before any conditions were evaluated.
We scored the full $282$-run T4 pool idempotently, with $14$ transient \texttt{claude -p} CLI errors retried until convergence (final $282/282$).
The rubric and the judge code are kept in the paper repository.

\begin{figure}[h]
\centering
\includegraphics[width=0.85\linewidth]{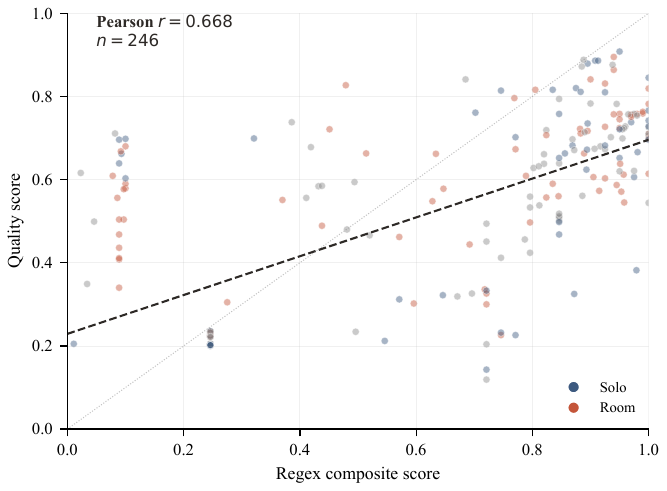}
\caption{The two scorers measure the same thing loosely, not interchangeably. Each point is one T4 run carrying both a regex composite and an LLM-judge composite ($n{=}246$, the runs scored by both); Pearson $r{=}0.67$. The spread off the diagonal is why the body uses one scorer throughout and reports the other as a cross-check rather than averaging them.}
\label{fig:judge_calibration}
\end{figure}

\paragraph{Cross-validation against regex scorer.}
We initially used a regex-based intrinsic scorer (described below) and rescored every run with the LLM-judge as a robustness check.
Pearson correlation over the $246$ T4 runs that carry both scores is $r{=}0.67$ (Figure~\ref{fig:judge_calibration}; the LLM-judge pool is $282$ runs and the $30$--$700$\,s analysis pool a $235$-run subset of it; the regex scorer covers a subsample of each); the LLM-judge spreads substrate-only conditions slightly more than the regex scorer and tightens the AgentRoom condition.
All main directional findings (CMH OR=$13.7$, AgentRoom $>$ parallel-merge with $p{<}0.01$, AgentRoom $>$ ChatDev with $p{<}0.01$, $2\times$Haiku Room $>$ $1\times$Sonnet Solo) hold under both scorers.
The regex scorer is retained below as a secondary scorer; with the AST scorer it makes up the regex/AST family behind the difficulty gradient, and the T5 matrix and the cross-model effect-size table use the AST scorer (Table~\ref{tab:crossmodel}).

\paragraph{Same-family judge concern.}
The LLM-judge here is Sonnet~4.6, evaluating outputs from a mix of distinct models including Haiku~4.5, Codex GPT-5.4, and Gemini~3~Flash, in addition to Sonnet itself.
Same-family preference~\citep{preference_leakage} can inflate Sonnet-vs-other comparisons; the cross-validation against the regex scorer at $r{=}0.67$ provides a partial guard, and the parallel-merge vs AgentRoom contrast (both Sonnet) is invariant to that bias because the judge sees the same producer model in both conditions.
A fully independent third-party judge is left for future work.

\paragraph{Regex scorer (secondary, historical).}
This paragraph documents the regex patterns and the rationale behind each dimension and weight.
Weights were fixed before any multi-agent runs were evaluated, based on the relative importance of each dimension for production-grade TypeScript code as judged by the authors and not tuned to favour any condition.

\begin{itemize}[leftmargin=*]
    \item \textbf{Type Safety} (weight 0.25): counts typed function parameters over total parameters; penalises \texttt{any} annotations (each occurrence subtracts a fixed fraction up to the dimension ceiling); rewards interface and type-alias definitions.
    Rationale: TypeScript's value over JavaScript is precisely the type surface; halving this weight would reward untyped escape hatches.
    \item \textbf{Defensive Coding} (weight 0.25): checks for input validation (\texttt{safeParse}, \texttt{status(400)}), 404 handling, error classes, \texttt{try/catch} blocks, null checks, and HTTP status-code diversity.
    Rationale: production code fails on edge cases; agents that skip validation score artificially high on happy-path tests alone.
    \item \textbf{Test Quality} (weight 0.20): counts test cases, assertions per test, error-path tests, and edge-case coverage via keyword search for \texttt{missing}, \texttt{invalid}, \texttt{duplicate}, \texttt{unauthorized}.
    Rationale: test count alone overrewards trivial tests; assertion density and error-path presence correct this.
    \item \textbf{Design} (weight 0.15): file separation ($\geq 3$ files), \texttt{const} over \texttt{let} preference, immutable patterns, meaningful comments, middleware usage.
    Rationale: design choices affect downstream maintainability but are weighted lower because they are partially captured by Type Safety and Defensive Coding already.
    \item \textbf{Test Pass Rate} (weight 0.15): fraction of tests passing, parsed from vitest's ``N passed | M failed'' summary line.
    Rationale: included at a moderate weight so that a fully non-compiling output scores near zero without letting a trivially small test suite dominate.
\end{itemize}

\paragraph{AST cross-check.} As a robustness check on our regex-based scorer, we implemented an independent AST-based scorer using the TypeScript compiler API (script at \texttt{scripts/ast\_scorer.mjs} in the paper repository).
The AST scorer counts language-level constructs (typed parameters, type alias and interface declarations, try/catch statements, throw statements, validation calls, test calls, function and class declarations) and combines them into a composite analogous to our regex scorer.
On a random sample of 30 runs spanning all tasks, models, and conditions, Pearson correlation between the two scorers is $r{=}0.79$ and Spearman rank correlation is $\rho{=}0.62$ ($n{=}30$).
The agreement is far from perfect (in particular the AST scorer treats Codex's compact single-file implementations more harshly than our regex scorer), but the strong correlation supports using the regex scorer as the cross-check in the cross-model ES comparisons; runs that the regex scorer rates highly are also rated highly by the AST scorer in the rank-order sense.
We do not retire the regex scorer in favor of the AST scorer because the regex scorer is more portable across runtime variants (ESM/CJS, partial syntax, half-finished files), but the cross-check addresses the concern that regex-only scoring could mistakenly reward lexical defensive-coding patterns disconnected from runtime behavior.

\paragraph{Leave-one-out sensitivity (alternative solo subset).} Table~\ref{tab:sensitivity} reports a robustness check on a smaller solo subset ($n{=}4$, mean 0.786) than the canonical T4 Sonnet solo used in main Table~\ref{tab:ablation} and Appendix~\ref{app:more_results}.
The robustness conclusion does not depend on the subset choice: the condition ordering (AgentRoom $>$ solo) is preserved under every single-dimension ablation, with ES ranging from 1.12 to 1.25 on this subset.

\begin{table}[h]
\centering
\caption{Scorer sensitivity: leave-one-out on T4 (Sonnet 4.6, solo vs $\times$2 AgentRoom). The solo mean reported here (0.786) is computed over a smaller subset (4 runs) than the canonical solo subset used in Table~\ref{tab:ablation}; the smaller subset is the intersection of solo runs for which we retained per-dimension intermediate scorer outputs at the time of writing, kept here only as a robustness check. The AgentRoom~$>$~solo ordering is preserved under every single-dimension ablation and the ratio stays within 1.12--1.25 on this subset.}
\label{tab:sensitivity}
\begin{tabular}{lccc}
\toprule
Removed dimension & Solo mean & $\times$2 mean & ES \\
\midrule
(none, full scorer) & 0.786 & 0.928 & 1.18 \\
Type Safety & 0.721 & 0.904 & 1.25 \\
Defensive Coding & 0.815 & 0.944 & 1.16 \\
Test Quality & 0.854 & 0.956 & 1.12 \\
Design & 0.793 & 0.921 & 1.16 \\
Test Pass Rate & 0.749 & 0.916 & 1.22 \\
\bottomrule
\end{tabular}
\end{table}

Regex-based scoring was chosen over AST parsing for reproducibility across runtime variants (ESM, CJS, mixed) and because agents occasionally emit near-valid code that an AST parser would reject outright; the trade-off is that the scorer is lexical and can be fooled by appearance of defensive code that is never exercised.
We treat this as a known limitation (\S\ref{sec:scoring}) and as a direction for future work.

\end{document}